# Mushroom image recognition and distance generation based on attention-mechanism model and genetic information

Wenbin Liao[2, #], Jiewen Xiao[1, #], Chengbo Zhao[1], Yonggong Han[2], ZhiJie Geng[2]

Jianxin Wang[2, *], Yihua Yang[4, *]

[1] College of Environmental Science and Engineering, Beijing Forestry University, 35 Qinghua East Road, Haidian District, Beijing 100083, P. R. China

[2] School of information Science and Technology, Beijing Forestry University, 35 Qinghua East Road, Haidian District, Beijing 100083, P. R. China

[3] Guizhou Institute of Biology， Guiyang National Economic and Technological Development Zone, Guizhou Province 550000, P. R. China

#: the contribution of authors is equal.

*Corresponding author:

Yihua Yang

E-mail: yangyh_009@163.com

Jianxin Wang

E-mail: wangjx@bjfu.edu.cn

**Abstract**

The species identification of Macrofungi, i.e. mushrooms, has always been a challenging task. There are still a large number of poisonous mushrooms that have not been found, which poses a risk to people's life. However, the traditional identification method requires a large number of experts with knowledge in the field of taxonomy for manual identification, it is not only inefficient but also consumes a lot of manpower and capital costs. In this paper, we propose a new model based on attention-mechanism, MushroomNet, which applies the lightweight network MobileNetV3 as the backbone model, combined with the attention structure proposed by us, and has achieved excellent performance in the mushroom recognition task. On the public dataset, the test accuracy of the MushroomNet model has reached 83.9%, and on the local dataset, the test accuracy has reached 77.4%. The proposed attention mechanisms well focused attention on the bodies of mushroom image for mixed channel attention and the attention heat maps visualized by Grad-CAM. Further, in this study, genetic distance was added to the mushroom image recognition task, the genetic distance was used as the representation space, and the genetic distance between each pair of mushroom species in the dataset was used as the embedding of the genetic distance representation space, so as to predict the image distance and species. identify. We found that using the MES activation function can predict the genetic distance of mushrooms very well, but the accuracy is lower than that of SoftMax. The proposed MushroomNet was demonstrated it shows great potential for automatic and online mushroom image and the proposed automatic procedure would assist and be a reference to traditional mushroom classification.

# 1. Introduction

Mushroom has been used as a product for a long time because of its unique flavor and rich protein, due to its high quality of proteins, polysaccharides, unsaturated fatty acids, mineral substances, triterpenes sterols, and secondary metabolites, mushrooms have always been appreciated for their vital role in protecting and curing various health problems (Ma et al., 2018a), meanwhile, mushrooms are also known as mycoremediation tool because of they rely on the efficient enzymes to remediation of different types of pollutants (Kulshreshtha et al., 2014). It is of great help to human production and life.

Mushroom is of great significance to human health, and its potential application value is also worthy of our development and utilization. However, there are about 20,000 known mushroom species in the world, including many mushrooms beneficial to human health. But, beyond the known range, there are still many poisonous mushrooms, which pose a threat to people's production and life. From 1959 to 2002, A total of 28,018 mushroom poisoning incidents have occurred. Due to the low efficiency and high cost of traditional mushroom identification method, it can not be widely used. Therefore, we need a new method with high efficiency and fast identification of mushrooms.

With the development of deep learning, we have gradually changed from algorithm driven to data-driven in the field of image recognition. By collecting a large number of training data, we carry out iterative training on the model to make the model learn the feature representation of the image. The more data, the higher the fitting degree of the model to the feature space. In computer vision, convolutional neural network is often used for image classification. For example, the test accuracy of LeNet-5 (Lecun & Bottou, 1998) on MNIST dataset is 98%, AlexNet (Krizhevsky et al., 2017) won the championship in ILSVRC2012, with top-1 and top-5 error rates of 37.5% and 17.0%, VGG (Simonyan & Zisserman, 2015) proves that the network depth can affect the performance of the model to a certain extent. ResNet (He et al., 2016) won the championship of ILSVRC2015 and proposed a shortcut connection for the degradation phenomenon, which greatly eliminated the difficult of neural network training due to excessive depth. With the proposal of attention mechanism (Vaswani et al., 2017), more and more studies use attention mechanism to improve the performance of the model. Attention mechanism can make the model pay more attention to the useful information in the image, and then make a more accurate classification.

So far, a large of research apply the deep learning in computer vision and make many achievements in the classification task. For example, (Islam et al., 2019) first proposes a Machine Learning approach to identify Bangladeshi birds according to their species, has made great progress by using VGG-16 network as model to extract the features from bird images. (Sajedi et al., 2019) try to recognition Actinobacterial strains by Machine learning methods, employ CNN as the feature extractor improved the accuracy about 4%, and finally reached 91.90% accuracy on the subclasses of UTMC.V2.DB. (Hassan & Maji, 2021) implemented auto-encoder and auto-encoder with convolutional neural network (AE-CNN) for plant identification from leaf, and experimental result demonstrates that more accurate results can be obtained in classification using AE-CNN which have an accuracy of 96%. Based on the Inception V3 architecture, (Rizwan Iqbal & Hakim, 2022) use various types of data enhancement methods to achieve the accuracy of mango classification of 99.2%. Furthermore, insect pest classification is a difficult task because of various kinds, scales, shapes, complex backgrounds in the field, and high appearance similarity among insect species. but (Ung et al., 2021) present different convolutional neural network-based models in this work, including attention, feature pyramid, and fine-grained models. the experimental results show that these models made great progress and have high accuracy.

In conclusion, it is a good choice to use convolutional neural network in mushroom recognition task. However, due to the large variety of mushrooms and the strong interference of the environment, the direct use of convolutional neural network can not achieve good recognition effect. Therefore, through the selection of attention mechanism and different attention structures, we embed them into the lightweight convolution neural network to construct the mushroom recognition model, that is, MushroomNet. We further study the model by add the genetic distance as the representation space and using the genetic distance between species as the embedding of image distance as a new image distance generation method, the model has good performance in predicting species genetic distance and species identification. Due to the small amount of parameters of skeleton structure and less computational resources, it is suitable for mobile terminal and is the best choice for field researchers.

## 2. Materials and methods

### 2.1. Image dataset

#### 2.1.1. Local dataset

In this study, we totally obtained 3377 mushroom images from different online resources, and then, all the mushroom images has been identified by professional experts of the taxonomic field and saved in PNG format for subsequent experiments. After analysis and identification, there are 18 species of mushrooms, which are: *Amanita*, *Armillaria mellea*, *Boletus*, *Cantharellus cibarius*, *Collybia*, *Ganoderma*, *Laccaria*, *Lactarius*, *Lentinus edodes*, *Morchella*, *Ophiocordyceps sinensis*, *Pleurotus citrinopileatus*, *Ramaria*, *Russula*, *Sarcodon imbricatum*, *Thelephora ganbajun*, *Trichloma matsutake* and *Tuber*, the sample figures shown as Figure 1.

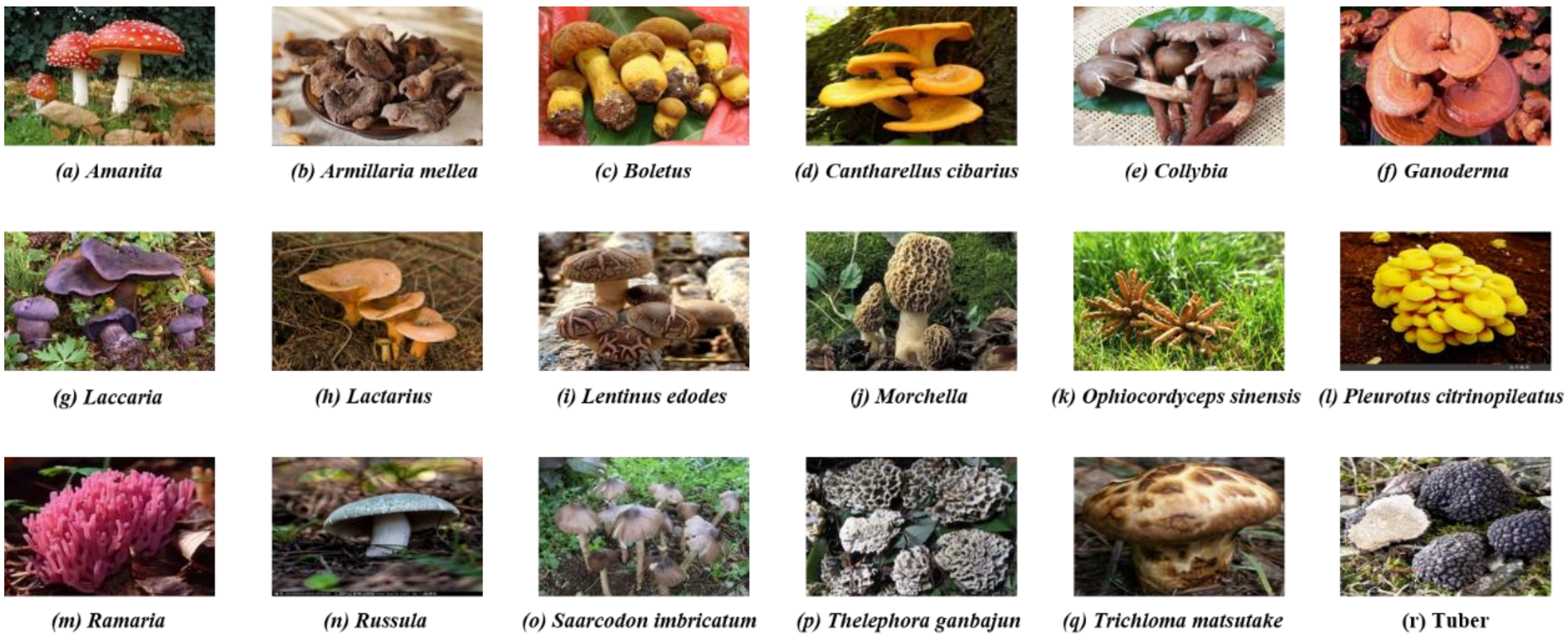

**Figure 1.** Sample images of local mushroom species

#### 2.1.2. Open dataset

We downloaded a public dataset from the Kaggle platform for subsequent exploration of the model we proposed, especially in terms of generalization ability. Kaggle is an open-source platform for data sharing and analysis. The public dataset we selected have 9528 images of 12 species of mushrooms. All mushroom images have been pre-processed to remove the wrong data, and then, identified and saved in different folders.

there are 12 species of mushrooms, which are: *Agaricus*, *Amanita*, *Boletus*, *Cortinarius*, *Entoloma*, *Exidia*, *Hygrocybe*, *Inocybe*, *Lactarius*, *Pluteus*, *Russula* and *Suillus*, shown as Figure 2.

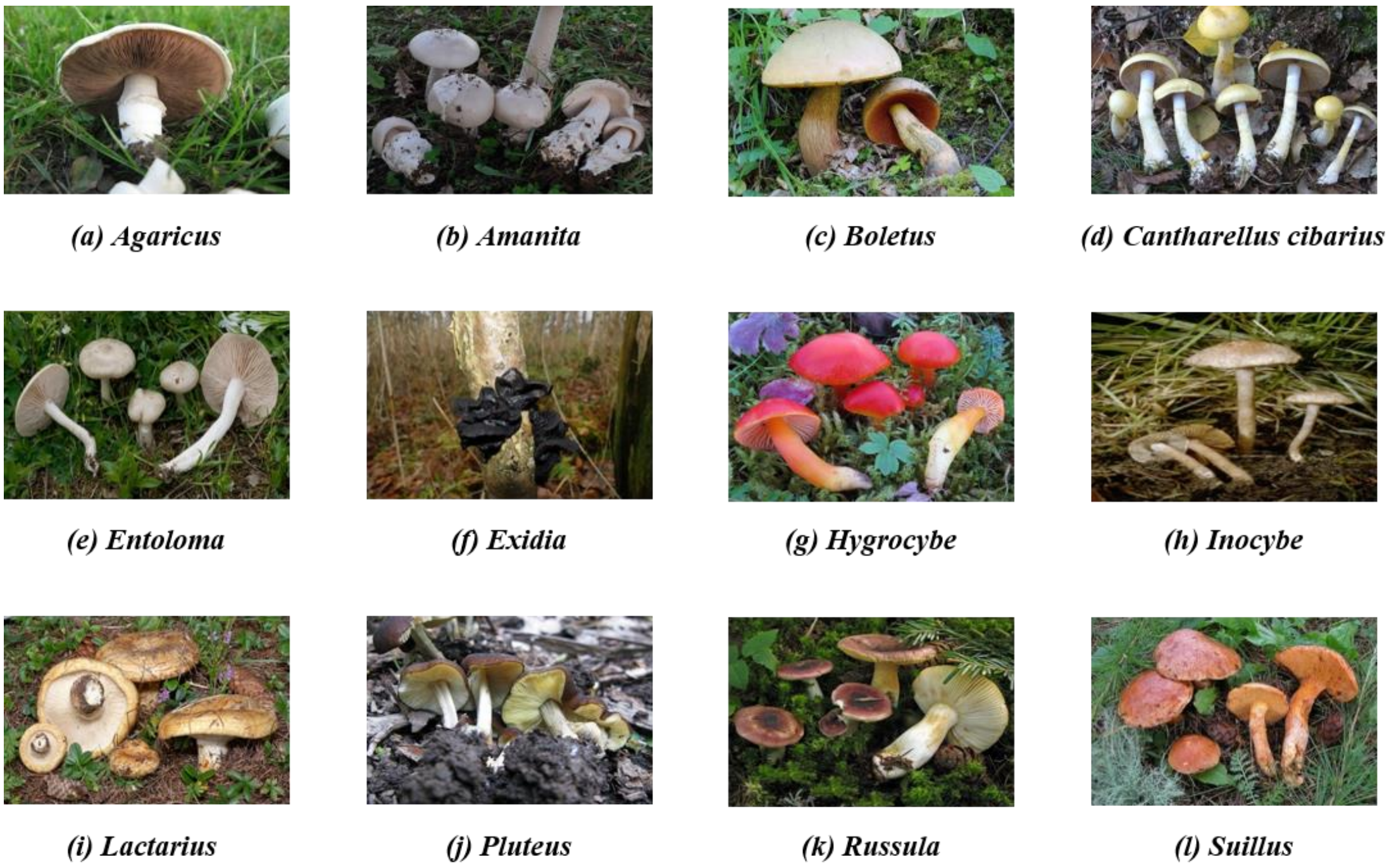

**Figure 2.** Sample images of open mushroom species

## 2.2. Overall process

In this paper, the whole process is divided into three stages: data preprocessing, model training and result evaluation, the flowchart is shown in Figure 3.

Firstly, the mushroom dataset is divided into three parts: training set, validation set and test set, with the proportions of 0.8, 0.1 and 0.1 respectively. The training set is used to fit the model, the validation set is used to adjust the hyperparameters of the model and preliminarily evaluate the performance of the model, and the test set is used to finally validate the generalization ability of the model. After that, in order to balance the number of samples of each mushroom species and prevent overfitting of the model, the data augmentation techniques are applied to the mushroom dataset to generate new images, which include image rotation, image clipping, image sharpening, image contrast, image brightness and etc. Then, we resize all images to $224 \times 224 \times 3$ as input of the model.

Secondly, the proposed model MushroomNet combined the advantage of the lightweight MobileNet-V3 (Howard et al., 2019) and the attention mechanism which contained channel attention (Hu et al., 2020) and ECANet (Wang et al., 2020), and then, model training was performed, in the training process, we use the validation set to evaluate the classification performance of the model, obtain the optimum model, and make the final evaluation on the test set.

Thirdly, now we get the optimum model, in order to understand the performance and generalization ability of the model more objectively, we use accuracy, precision, F1 and recall to evaluate the model. Then, in order to understand the interpretability of the model, we use the output of the last convolution layer of the model to draw the heat map and overlay the heat map on the original image.

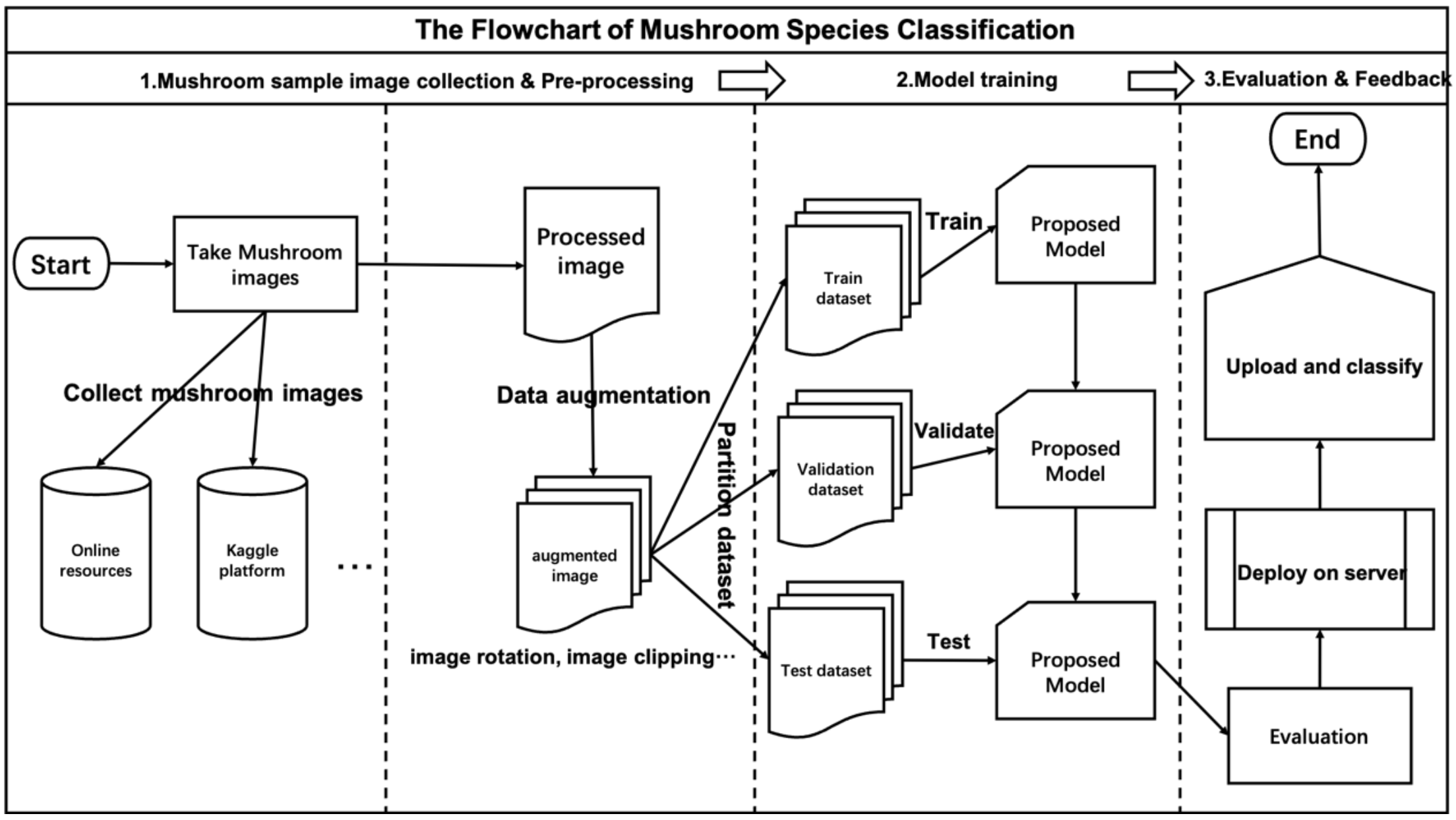

**Figure 3.** The flowchart of mushroom species classification

## 2.3. Related work

### 2.3.1. Attention mechanism

In the process of information processing, people often pay attention to some information and ignore some unimportant information, which is called attention mechanism (Vaswani et al., 2017). In computer vision, attention mechanism has achieved good performance, such as channel attention (CAM), spatial attention (SAM) (Zhu et al., 2020). In this paper, we apply the channel attention (CAM) and efficient channel attention (ECANet) to the proposed model.

Channel attention can be divided into two operations: Squeeze and Excitation. First step, use a global average pooling operation to squeeze a feature map with dimension $C \times H \times W$ into a feature map with dimension $C \times 1 \times 1$, and then use two full connection layers on the feature map. After the first full connection layer, the feature map dimension is reduced to $C/r$, and after the second full connection layer, the feature map dimension is changed back to $C$, Where $r$ is a hyperparameter. It is proved by experiments that $r$ is generally 16, and the specific mathematical formula is shown in Eqs. (1-2):

$$Z_c = F_{sq}(u_c) = \frac{1}{H \times W} \sum_{i=1}^{H} \sum_{j=1}^{W} u_c(i,j) \tag{1}$$

$$s = F_{ex}(z, W) = \sigma\big(g(z, W)\big) = \sigma\big(W_2 \delta(W_1 z)\big) \tag{2}$$

ECANet is an improvement of CAM, which uses one-dimensional convolution instead of the full connection layer to avoid the possible loss of information in dimension reduction. Its mathematical formula is shown in Eq. (3):

$$F(W_1, W_2, y) = W_2 ReLu(W_1 y) \tag{3}$$

### 2.3.2. MobileNet-V3

MobileNet-V3 is the latest version of MobileNet series. It integrates the deep-wise separable convolution of MobileNet-V1 (Howard et al., 2017) and the inverted residual block and linear bottleneck structures of MobileNet-V2 (Sandler et al., 2018), which uses NAS (neural structure search) (Zoph & Le, 2017) to search the network configuration and parameters.

The MobileNet-V3 advances the average pooling layer. It uses $1 \times 1$ convolutional operation to expand, the pool layer activation function is immediately followed, and finally output by 1x1 convolutional operation.

In MobileNet-V2, there is a 1x1 convolution layer before avg pooling, which aims to improve the dimension of the feature map and is more conducive to the prediction of the structure. However, this brings a certain amount of computation, so it is placed behind avg pooling. Firstly, avg pooling is used to reduce the size of the feature map from 7x7 to 1x1, and then a 1x1 convolution kernel is used to improve the dimension, this reduces the amount of calculation by 49 times. to further, reduce the amount of calculation, the previous 3x3 and 1x1 convolution are removed to further reduce the amount of calculation, and the accuracy is not greatly lost.

### 2.3.3. Transfer learning

The machine learning algorithm (Pan, 2011) based on neural network needs a large amount of labeled data, and labeling a large amount of data is a very challenging task. So, migration learning is a common scheme in neural network. It is to migrate the trained model parameters on large dataset to new model to help the new model train on small dataset, because most of the data or tasks are relevant, we can share the learned model parameters, which can also be understood as the knowledge learned by the model, with the new model in some way, so as to speed up and optimize the learning efficiency of the model without learning from zero like most networks, the flow of transfer learning is shown as Figure 4.

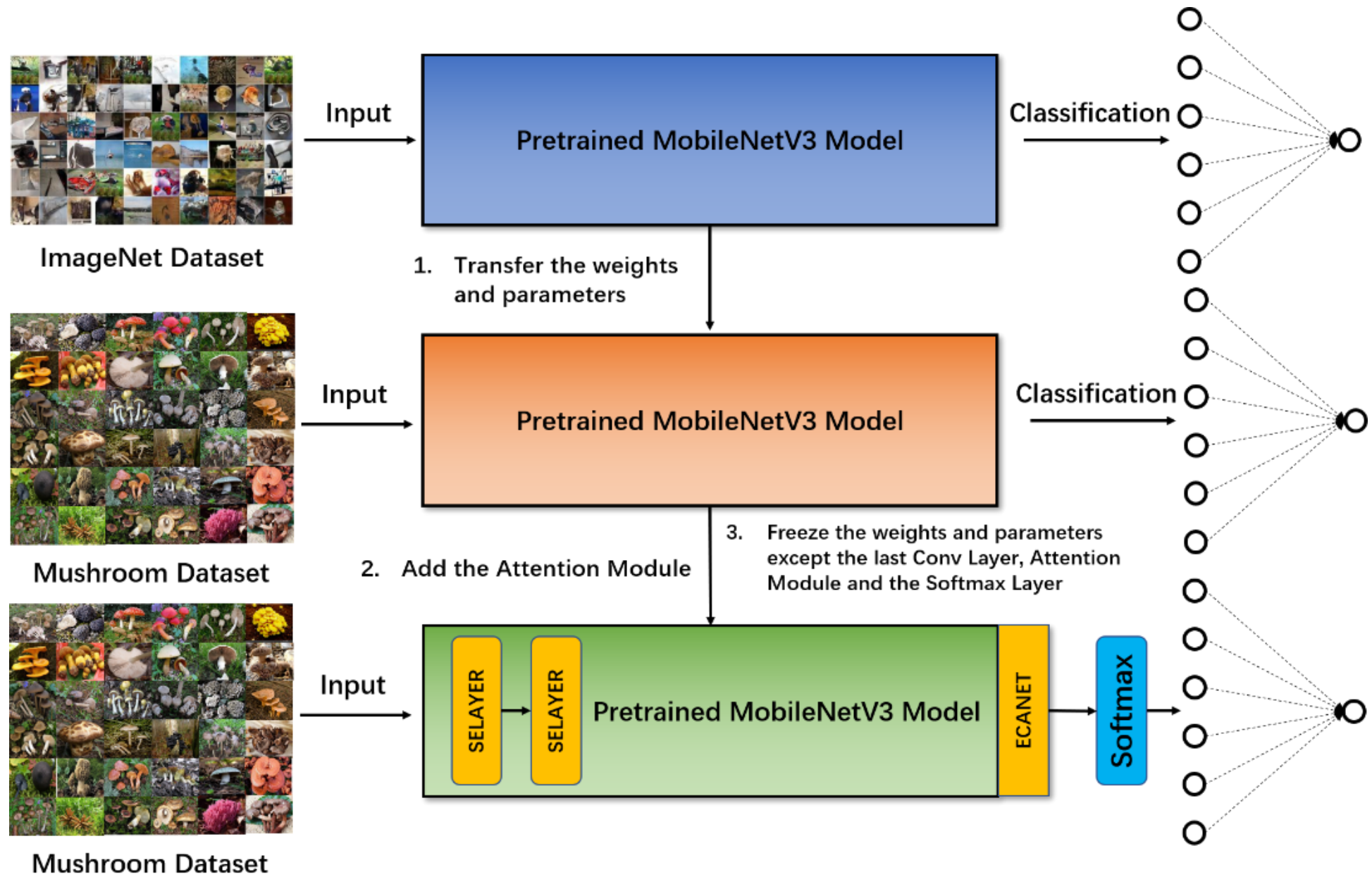

**Figure 4.** The flowchart of transfer learning

### 2.4. Proposed approach

#### 2.4.1. MushroomNet

As the latest version of mobilenet series, MobileNet-V3 integrates the deep separable convolution of MobileNet-V1, the inverted residuals of MobileNet-v2 and the linear bottleneck, and uses h-swish instead of swish function. Compared with MobileNet-V2, the accuracy is improved by about 3.2%, but the time is reduced by 15%. It is the best choice for mobile end model. In addition, because the attention mechanism can well understand the importance of data and is widely used in computer vision, we use Mobilenet-V3 to combine channel attention and ECANet to propose our model MushroomNet. MushroomNet abandons the classification layer at the tail of Mobilenet-V3 and adds two channel attention modules behind the first convolution layer of the pre-trained model, Used for high-dimensional feature extraction. Next, add ECANet after the last convolution layer of Mobilenet-V3, and finally add a fully connected softmax as the classification layer for the classification and identification of mushroom species. The network structure is shown in Figure 5, and the network structure parameters are shown in Table 1.

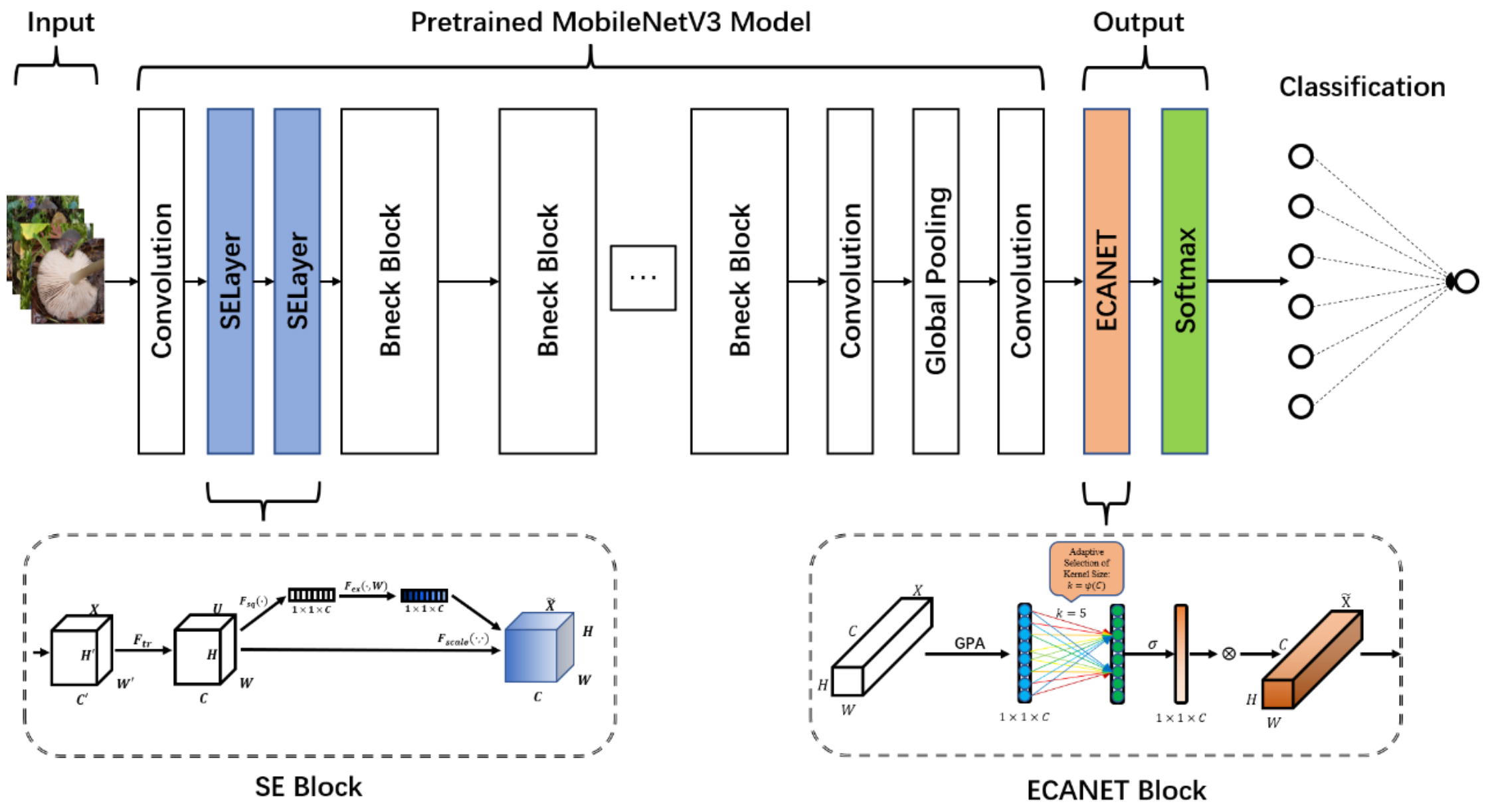

**Figure 5.** MushroomNet Structure

**Table 1.** Parameters of MushroomNet

| Module | Input | exp size | out | SE | NL | s |
|---|---|---|---|---|---|---|
| conv2d | 224 × 224 × 3 | - | 16 | - | HS | 2 |
| SE | 224 × 224 × 16 | - | - | - | - | - |
| SE | 224 × 224 × 16 | - | - | - | - | - |
| bneck,3x3 | 122 × 112 × 16 | 6 | 16 | - | RE | 1 |
| bneck,3x3 | 122 × 112 × 16 | 64 | 24 | - | RE | 2 |
| bneck,3x3 | 56 × 56 × 24 | 72 | 24 | - | RE | 1 |
| bneck,5x5 | 56 × 56 × 24 | 72 | 40 | P | RE | 2 |
| bneck,5x5 | 28 × 28 × 40 | 120 | 40 | P | RE | 1 |
| bneck,5x5 | 28 × 28 × 40 | 120 | 40 | P | RE | 1 |
| bneck,3x3 | 28 × 28 × 40 | 240 | 80 | - | HS | 2 |
| bneck,3x3 | 14 × 14 × 80 | 200 | 80 | - | HS | 1 |
| bneck,3x3 | 14 × 14 × 80 | 184 | 80 | - | HS | 1 |
| bneck,3x3 | 14 × 14 × 80 | 184 | 80 | - | HS | 1 |
| bneck,3x3 | 14 × 14 × 80 | 480 | 112 | P | HS | 1 |
| bneck,3x3 | 14 × 14 × 112 | 672 | 112 | P | HS | 1 |
| bneck,5x5 | 14 × 14 × 112 | 672 | 160 | P | HS | 2 |
| bneck,5x5 | 7 × 7 × 160 | 960 | 160 | P | HS | 1 |
| bneck,5x5 | 7 × 7 × 160 | 960 | 160 | P | HS | 1 |
| conv2d,1x1 | 7 × 7 × 160 | - | 960 | - | HS | 1 |
| pool,7x7 | 7 × 7 × 960 | - | - | - | - | 1 |
| conv2d,1x1, NBN | 1 × 1 × 960 | - | 1280 | - | HS | 1 |
| ECANET | 1 × 1 × 1280 | - | - | - | - | - |
| Softmax | 1 × 1 × 1280 | - | k | - | - | - |

#### 2.4.2. Training strategy

The model training stage is divided into three parts. In the first stage, we initialize the MobileNet-V3 model and load the parameters and weights of MobileNet-v3 pre trained from ImageNet through transfer learning. In the second stage, the classification layer of the model is replaced by the classification layer suitable for our mushroom dataset, and the model is trained on our dataset to fine tune the parameters and weights and retrain the classification layer. In the third stage, we add the attention modules to the model obtained from the second stage, then freeze other layers of the model except the last convolution layer and softmax layer to prevent changing the trained parameters and weights, and finally train the attention modules from scratch on our dataset.

In the first stage, the transfer learning technology is adopted. Through training on the ImageNet dataset, the parameters and weights are loaded into the mobilenetv3 model to initialize the model.

In the second stage, to fit our dataset, we replaced the classification layer, use Adam optimizer to train and update the parameters and weights, and set the learning rate as $1\times10^{-4}$.

In the third stage, we added the attention modules, frozen other layers of the model except the last convolution layer and the softmax layer, trained the attention modules with our dataset, updated the weight with Adam optimizer, and set the learning rate as $1\times10^{-4}$.

## 2.5. A new method for image distance generation based on genetic distance and its application in species identification and distance prediction

### 2.5.1. Calculation of the genetic distance of the dataset

The calculation of genetic distance within this study was performed using the DNA sequence of the ITS region. The ITS sequences of the web-sourced dataset were derived from NCBI (https://www.ncbi.nlm.nih.gov/nuccore). The DNA sequence of the local dataset is shown in Table S1, and the DNA data of the public dataset is shown in Table S2. The sequences were aligned using Clustal W of MEGA11, the base alignment Gap opening penalty was 15, the Gap expansion penalty was 6.66, the multiple sequence alignment Gap opening penalty was 15, and the Gap expansion penalty was 6.66. The DNA matrix weight is IUB and the exchange weight is 0.5. After alignment, the genetic distance was calculated using Pairwise Disantance, and the uncertainty was calculated using 100 Bootstrap samples. The replacement model uses Nucleotide, calculated using the Maximum Composite Likelihode method. After the calculation, the lower triangular matrix is derived, the diagonal zeros are filled, and the upper triangular matrix is generated by transposing, which is merged into a genetic distance square matrix.

### 2.5.2. Phylogenetic analysis

The aligned sequences were constructed using the Maximum Likelihood Tree method to construct a phylogenetic tree, the phylogenetic test was sampled 100 times using the Bootstrap method, the replacement model was replaced by Nucleotide, and the Tamura-Nei model was used for calculation.

### 2.5.3. Image distance generation based on genetic distance

In this study, the genetic distance was used as the representation space, and the genetic distance between the mushroom species in the dataset was used as the embedding of the genetic distance representation space. This study uses a convolutional neural network to extract image features and map the image features to the feature space of a fully connected layer (Fc). The feature space and representation space are connected by softmax or MSE function, that is, the Embedding of genetic distance. Through learning on the training set, the model acquires the ability to generate genetic distances from image features, that is, to obtain image distances with biological genetic information through feature extraction.

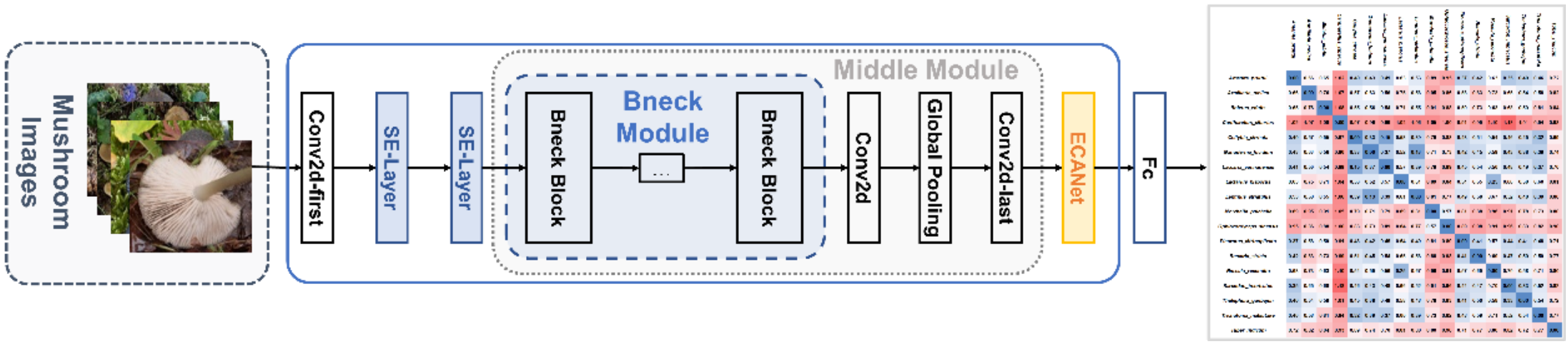

**Figure 6** Image distance generation based on genetic distance

### 2.5.4. Application of genetic distance for mushroom identification

In this study, the distance calculation is performed between the Embedding of the genetic distance predicted by the model and the ground truth of the genetic distance, including cosine distance, Euclidean distance, etc. The obtained distance is the Label Embedding, which can be mapped to the Label space through the Label Embedding ( Label Sapce) to complete mushroom image classification.

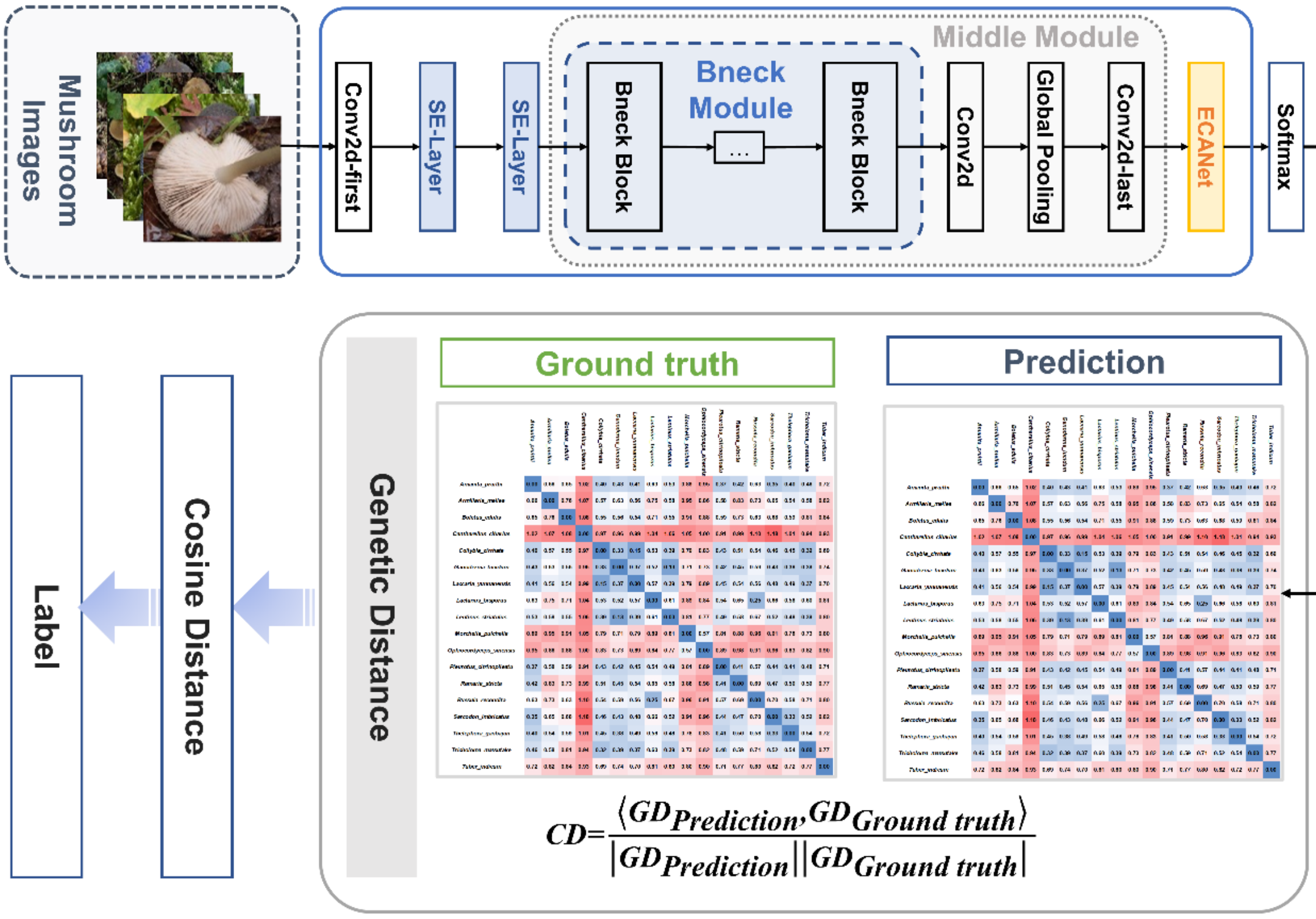

**Figure 7** Image recognition based on genetic distance

### 2.5.5. Improved genetic distance for mushroom recognition

Based on the research results, this study found that Softmax would increase the distance between different pictures in the feature space, and the MES function would have a large uncertainty on the diagonal elements. Therefore, this study improved the diagonal elements of genetic distance, Make the diagonal genetic distance -1, perform model training, and perform cosine distance calculation with the ground truth whose diagonal is 0 to obtain label embedding.

## 3. Experimental results and analysis

In our experiments, except for some image pre-processing work performed by the Photoshop tool, the main algorithms such as data augmentation and CNNs were conducted using Anaconda3 (Python 3.6), PyTorch-GPU library, and OpenCV-python3 library, etc. The training and testing of CNNs were accelerated by GPU, and the experimental hardware configuration included Intel® Xeon(R) E5-2620 CPU (2.10 GHz), 64-GB memory, and NVIDIA GeForce RTX 2080 (CUDA 10.2) graphics card, which was used for the model training and testing.

### 3.1. Experiments on our local dataset

According to the training strategy described in Section 2.4.2, we divide the local dataset into training set, validation set and test set, with the proportion of 0.8, 0.1 and 0.1. Use transfer learning technique to load the parameters and weights trained from ImageNet for the model and initialize the model, then, remove the final classification layer and replace it with a new classification layer suitable for our local dataset, We use the dataLoader provided by PyTorch to load the data and shuffle the order of the data to avoid overfitting. We set the learning rate as 1e-4 and the optimizer as Adam. Then we use a batch of 12 images for a total of 30 epochs for the second stage of model training. Next, load the model parameters and weights after second stage of model training and perform the third stage of training in combination with the attention modules. Note that the parameters and weights of other layers except the classification layer, softmax layer and attention module need to be frozen in the third stage of training to prevent the training from changing the trained parameters and weights information. In order to compare the performance differences between the models, in the experiment, we selected nine classical models commonly used in computer vision: VGG, Squeezenet (Iandola et al., 2016), Xception (Chollet, 2017), Densenet121 (Huang et al., 2017), Resnet50 (He et al., 2016), ShuffleNet-V2 (Ma et al., 2018b), MobileNet-V2, MobileNet-V3 and EfficientNet-B0 (Tan & Le, 2019), which also use the parameters and weights information trained by transfer learning from ImageNet, Use the classification layer suitable for our local dataset to replace the original top layer, and then conduct retraining on our local dataset. Compare and analyze the results with the model we proposed. Each experiment is repeated in multiple groups to ensure the authenticity and effectiveness of the data.

The experimental results are shown in Table 2. Compared with the other nine models, after 30 epochs training, the accuracy of our proposed model in the validation set has achieved 73.8%, which is only a little lower than EfficientNet-B0. On the test set, our proposed model has the top performance of all models, the accuracy has achieved 77.4%, and the loss is relatively low compared with other models, in addition, because our model occupies a small memory and has fast execute speed, it is very suitable for mushroom recognition in mobile end.

**Table 2.** Accuracy and loss of different models on local dataset. [a]

| Models | Val accuracy | Val loss | Test accuracy | Test loss |
|---|---|---|---|---|
| VGG | 48.9% | 3.52 | 50.1% | 3.56 |
| Squeezenet | 52.2% | 2.81 | 58.9% | 2.33 |
| Xception | 68.9% | 1.64 | 71.1% | 1.48 |
| **Densenet121** | **76.4%** | **1.16** | **76.3%** | **1.16** |
| Resnet50 | 69.9% | 1.49 | 66.5% | 1.70 |
| Shufflenetv2 | 71.5% | 1.29 | 75.1% | 1.21 |
| Mobilenetv2 | 71.9% | 1.51 | 72.9% | 1.34 |
| Mobilenetv3 | 72.1% | 1.81 | 72.9% | 1.36 |
| **Efficientnet_B0** | **75.4%** | **1.27** | **76.9%** | **1.08** |
| **Proposed Model** | **73.8%** | **1.95** | **77.4%** | **1.72** |

[a] The models were trained for 30 epochs.

For classification problems, accuracy, precision, F1 and recall are generally used to measure the performance of the model, the relevant calculation formula is shown in Eqs. (4-7):

$$Accuracy = \frac{TP + TN}{TP + TN + FP + FN} \quad (4)$$

$$Precision = \frac{TP}{TP + FP} \quad (5)$$

$$Recall = \frac{TP}{TP + FN} \quad (6)$$

$$F1 = 2 \times \frac{Precision \times Recall}{Precision + Recall} \quad (7)$$

Where the correct classification also called true positive (TP) and true negative (TN), and misclassification also called false positive (FP) and false negative (FN), which can be obtained by statistical analysis of the data given by the model,

It can be seen from the ROC curve in Figure 6 that the TPR (true positive rate) of 18 species of mushrooms is very high, while the FPR (false positive rate) is low. The results show that our proposed model can well distinguish the features of different mushrooms and classify them. The confusion matrix in Figure 7 describes the number of correct and wrong classifications of different mushroom images. It can be seen that most mushrooms are correctly classified. However, due to the complexity of mushroom shooting environment and the similarity of mushrooms themselves, very few mushrooms are misclassified, such as mushroom *Ophiocordyceps sinensis*, there are 29 images in its test sample, 24 of which are correctly classified, and 5 of which are misclassified. The classification accuracy of this mushroom species is 80.00%, the precision is 96.00%, the F1 is 88.89%, and the recall rate is 82.76%. The detailed classification metrics of other mushrooms are shown in Table 3.

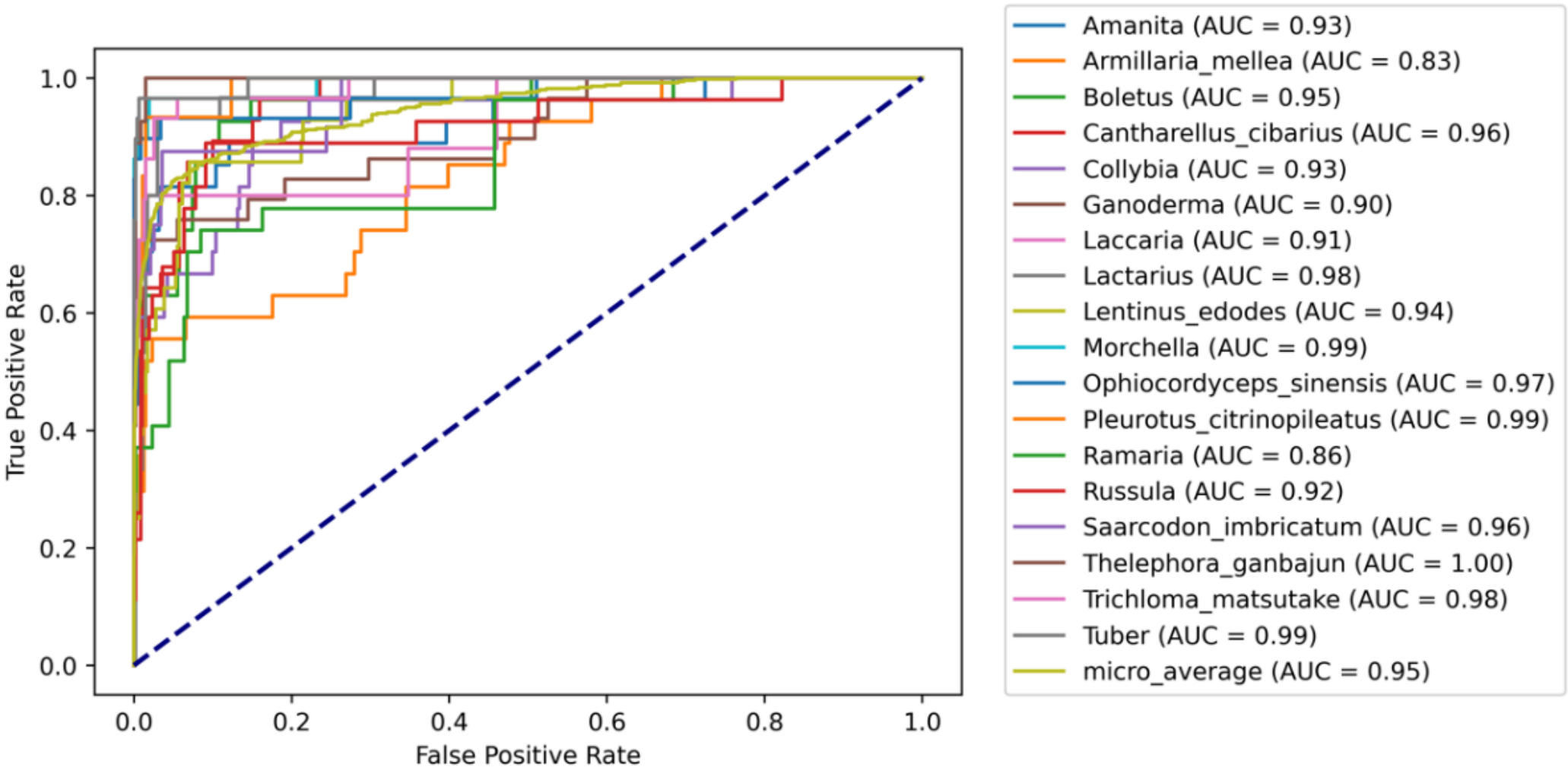

**Figure 6.** ROC curve of local dataset

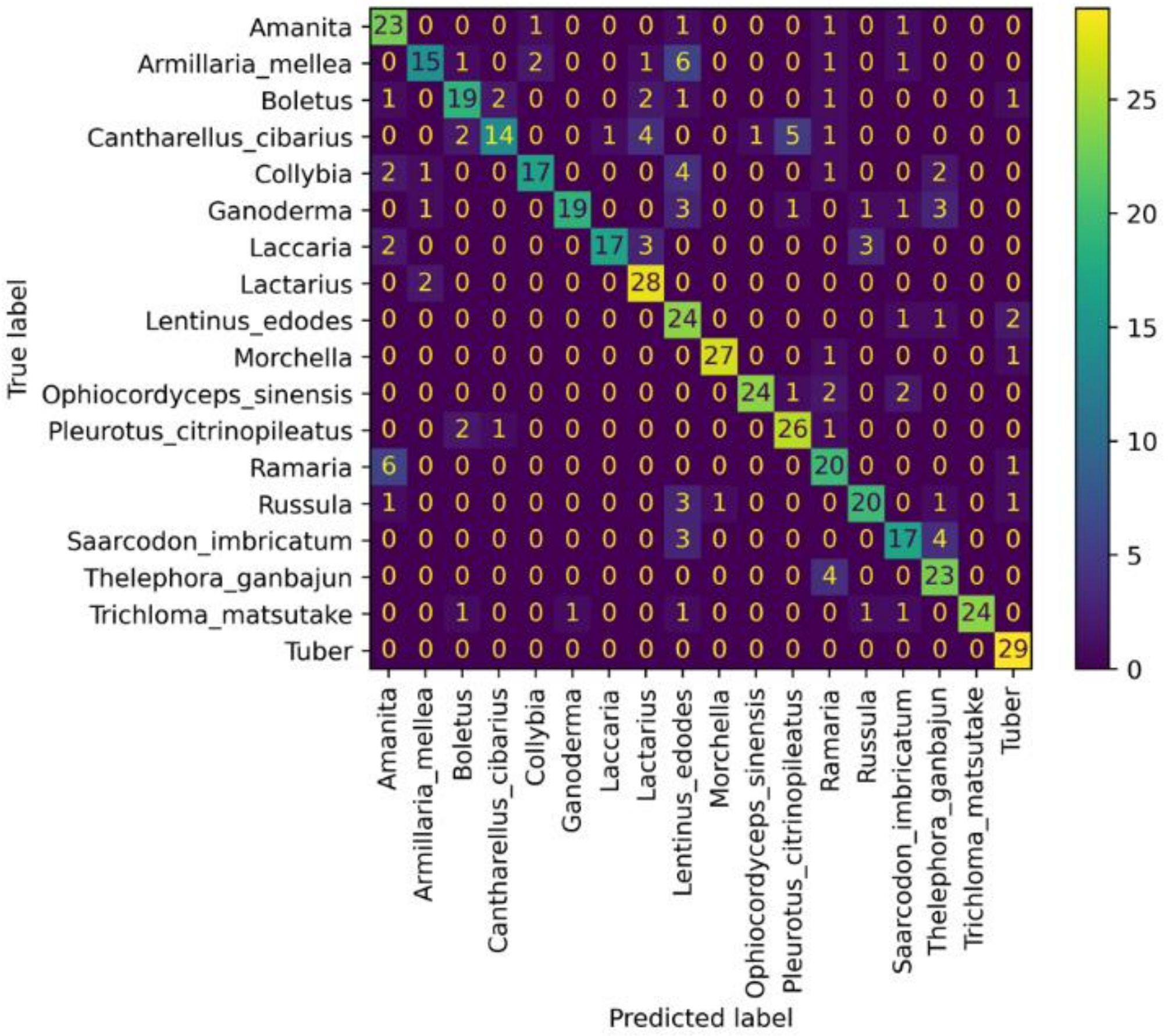

**Figure 7.** Confusion matrix of local dataset

**Table 3.** Classification metrics of different mushroom species

| ID | Classes | Identified samples | Correct samples | Accuracy (%) | Precision (%) | F1 (%) | Recall (%) |
|---|---|---|---|---|---|---|---|
| 1 | *Amanita* | 27 | 23 | 58.97 | 65.71 | 74.19 | 85.19 |
| 2 | *Armillaria mellea* | 27 | 15 | 48.39 | 78.95 | 65.22 | 55.56 |
| 3 | *Boletus* | 27 | 19 | 57.58 | 76.00 | 73.08 | 70.37 |
| 4 | *Cantharellus cibarius* | 28 | 14 | 45.16 | 82.35 | 62.22 | 50.00 |
| 5 | *Collybia* | 27 | 17 | 56.67 | 85.00 | 72.34 | 62.96 |
| 6 | *Ganoderma* | 29 | 19 | 63.33 | 95.00 | 77.55 | 65.52 |
| 7 | *Laccaria* | 25 | 17 | 65.38 | 94.44 | 79.07 | 68.00 |
| 8 | *Lactarius* | 30 | 28 | 70.00 | 73.68 | 82.35 | 93.33 |
| 9 | *Lentinus edodes* | 28 | 24 | 48.00 | 52.17 | 64.86 | 85.71 |
| 10 | *Morchella* | 29 | 27 | 90.00 | 96.43 | 94.74 | 93.10 |
| 11 | *Ophiocordyceps sinensis* | 29 | 24 | 80.00 | 96.00 | 88.89 | 82.76 |
| 12 | *Pleurotus citrinopileatus* | 30 | 26 | 70.27 | 78.79 | 82.54 | 86.67 |
| 13 | *Ramaria* | 27 | 20 | 50.00 | 60.61 | 66.67 | 74.07 |
| 14 | *Russula* | 27 | 20 | 62.50 | 80.00 | 76.92 | 74.07 |
| 15 | *Saarcodon imbricatum* | 24 | 17 | 54.84 | 70.83 | 70.83 | 70.83 |
| 16 | *Thelephora ganbajun* | 27 | 23 | 60.53 | 67.65 | 75.41 | 85.19 |
| 17 | *Trichloma matsutake* | 29 | 24 | 82.76 | 100.00 | 90.57 | 82.76 |
| 18 | *Tuber* | 29 | 29 | 82.86 | 82.86 | 90.63 | 100.00 |
| - | - | 499 | 386 | 77.35 | 77.35 | 77.35 | 77.35 |

In order to make better decisions on the model, we can use Grad-CAM technique to make visual interpretation and analysis of the model. Grad-CAM uses the gradient information of the last convolution layer flowing into model to assign important values to each neuron. It can visualize CNN networks with any structure without modifying the original network structure, as shown in Figure 8, The top image of (a) - (d) is the original test sample, and the bottom image is the image formed after the heat map is generated by Grad-CAM technique and superimposed with the original image. We can more intuitively understand the execute mechanism of the model in the whole classification process. From the image, we can clearly see that the model has successfully identified the boundary between the environment and mushrooms, most of the attention is focused on the features of mushrooms, which shows that the model indeed classifies mushrooms through the features of mushrooms. The model successfully learns the differences between different species of mushrooms. However, due to the complexity of the environment and the problem of photographing angle, there will be a small number of mushroom classification errors, but from the overall situation, The model still achieves the best effect in mushroom classification task.

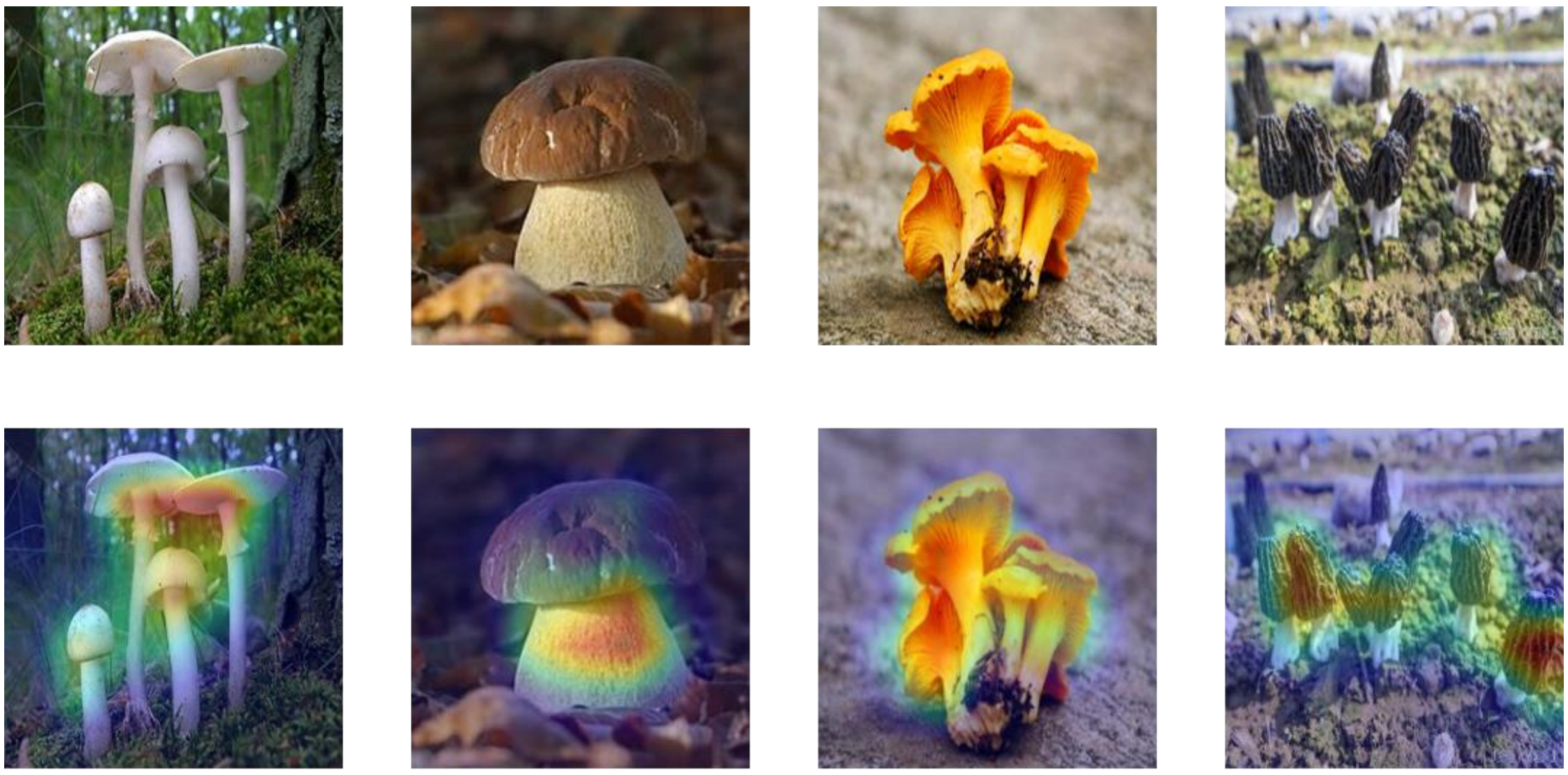

**(a)** **(b)** **(c)** **(d)**

**Figure 8.** Heat map of several mushroom species

### 3.2. Experiments on a public dataset

In order to validate the generalization ability of our proposed model, we downloaded a public mushroom dataset from the kaggle platform. The dataset contains 9528 mushroom images, a total of 12 species. We divided the data set into training set, validation set and test set, with the ratio of 0.8, 0.1 and 0.1. Through data enhancement techniques, such as image rotation, image sharpening, image clipping, image contrast and etc. to generate new images, ensure that the number of images of each species of mushroom is basically equal, and prevent the model from overfitting. Next, according to the training strategy described in 2.4.2, the model is trained in three stages. First, load the parameters and weight information of Mobilenet-V3 trained from ImageNet, and then remove the last classification layer and replace it with the classification layer that fit the public mushroom dataset, and retrain the model. Finally, add the attention modules, freeze the parameters and weights of other layers except the last convolution layer and softmax layer, and train the attention module separately. There are 12 images in each batch, a total of 30 epochs. In order to compare the performance of the model, use other models in 3.1 as the contrast, and carry out experiments for many times in each group to prevent contingency.

The experimental results are shown in Table 4. Our proposed model achieves the best performance in both the validation set and the test set, with an accuracy of 82.0% in the validation set and 83.9% in the test set. Compared with other models, the effect is significantly improved, indicating that our model has achieved good performance in different dataset and has strong generalization ability.

In addition, the ROC curve is shown in Figure 9. When the FPR (false positive rate) is low, the TPR (true positive rate) of most mushroom species is high. The experiment proves that the model successfully distinguishes different mushroom species and learns the small features of mushrooms. The results of confusion matrix also illustrate this. As shown in Figure 10, most test samples are correctly classified by the model. A few of them were misclassified due to external environment and other factors, such as the *Hygrocybe* with the test sample number of 112, and the correct classifying number reached 104. Eight images were misclassified, with an accuracy of 88.14%, an accuracy of 94.55%, an F1 of 93.69%, and a recall of 92.86%. The classification metrics of other mushroom species are shown in Table 5.

Through the heat map shown in Figure 11, we can intuitively see that our proposed model focuses on the main features of mushrooms and ignores the differences in the environment. The results show that in the complex field environment, our model can still distinguish the small features of mushrooms and classify them. In addition, due to the uncertainty of the field environment, there are a few misclassifications in the model, but this does not affect the performance of the model in the mushroom classification task. In the case of lack of relevant information on mushrooms in the field, the model we proposed is still the best choice.

**Table 4.** Accuracy and loss of different models on open dataset

| Models | Val accuracy | Val loss | Test accuracy | Test loss |
|---|---|---|---|---|
| VGG | 51.6% | 3.63 | 50.6% | 3.66 |
| Squeezenet | 62.4% | 1.72 | 65.8% | 1.54 |
| Xception | 80.6% | 0.88 | 80.9% | 0.91 |
| **Densenet121** | **79.2%** | **0.88** | **82.5%** | **0.82** |
| Resnet50 | 76.9% | 1.07 | 79.0% | 0.98 |
| Shufflenetv2 | 74.1% | 1.10 | 77.2% | 0.98 |
| Mobilenetv2 | 75.3% | 1.09 | 79.0% | 0.98 |
| **Mobilenetv3** | **79.2%** | **0.94** | **82.6%** | **0.86** |
| **Efficientnet_B0** | **81.7%** | **0.84** | **82.5%** | **0.85** |
| **Proposed Model** | **82.0%** | **1.10** | **83.9%** | **1.00** |

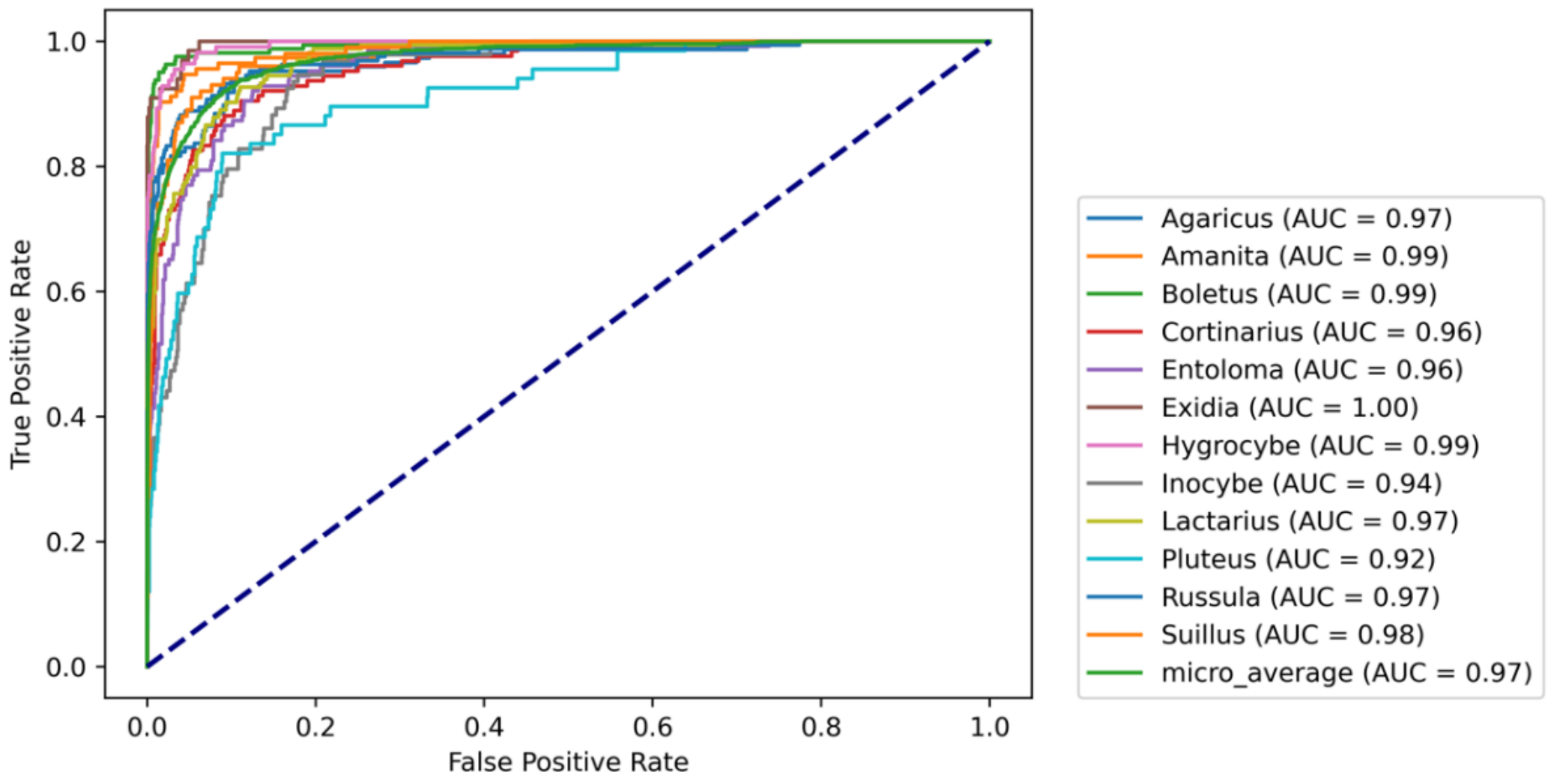

**Figure 9.** ROC curve of open dataset

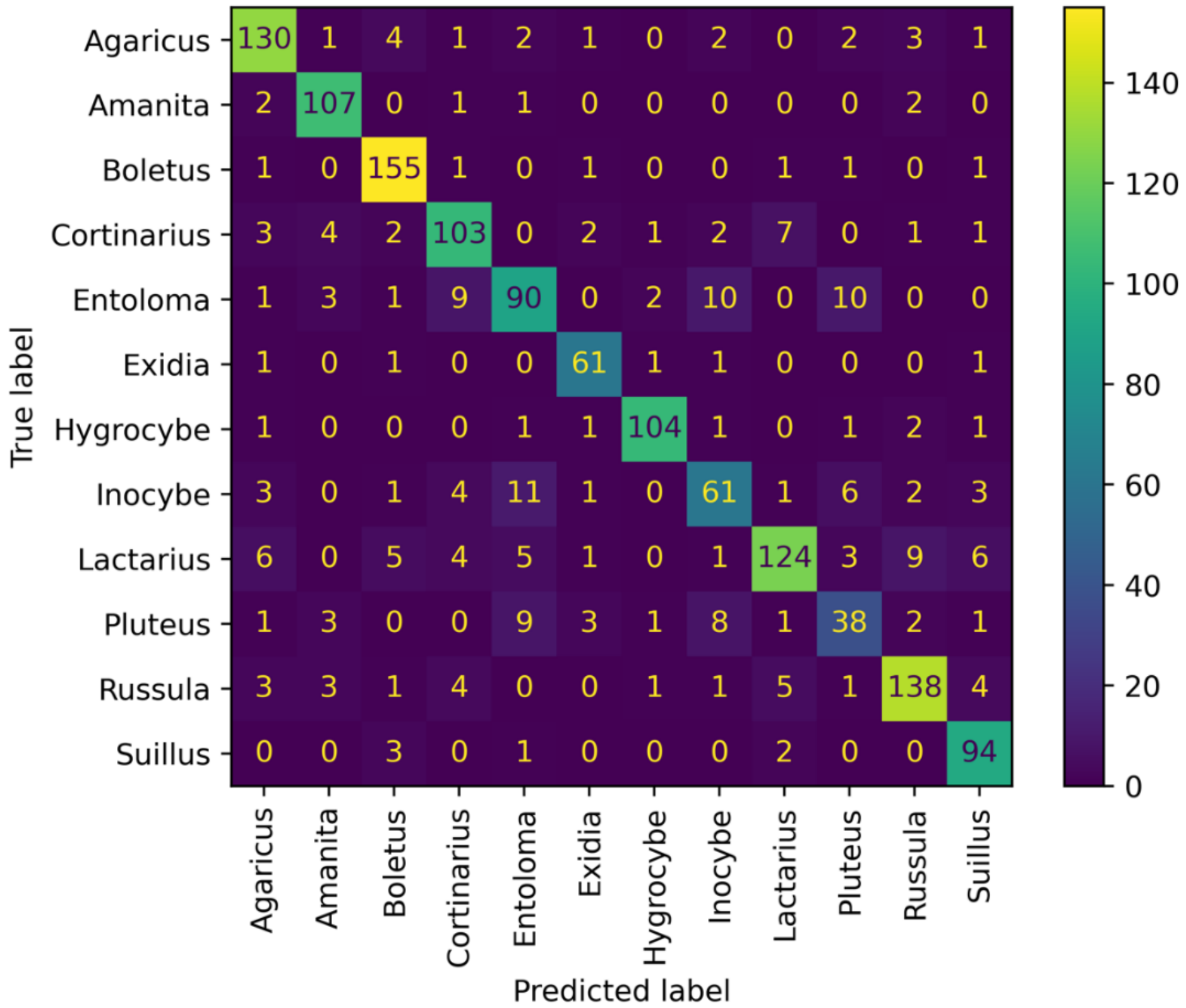

**Figure 10.** Confusion matrix of open dataset

**Table 5.** Classification metrics of different mushroom species

| ID | Classes | Identified samples | Correct samples | Accuracy (%) | Precision (%) | F1 (%) | Recall (%) |
|---|---|---|---|---|---|---|---|
| 1 | *Agaricus* | 147 | 130 | 76.92 | 85.53 | 86.86 | 88.44 |
| 2 | *Amanita* | 113 | 107 | 84.25 | 88.43 | 91.45 | 94.69 |
| 3 | *Boletus* | 161 | 155 | 86.59 | 89.60 | 92.81 | 96.27 |
| 4 | *Cortinarius* | 126 | 103 | 68.67 | 81.10 | 81.42 | 81.75 |
| 5 | *Entoloma* | 126 | 90 | 57.69 | 75.00 | 73.17 | 71.43 |
| 6 | *Exidia* | 66 | 61 | 80.26 | 85.92 | 89.05 | 92.42 |
| 7 | *Hygrocybe* | 112 | 104 | 88.14 | 94.55 | 93.69 | 92.86 |
| 8 | *Inocybe* | 93 | 61 | 51.26 | 70.11 | 67.78 | 65.59 |
| 9 | *Lactarius* | 164 | 124 | 68.51 | 87.93 | 81.31 | 75.61 |
| 10 | *Pluteus* | 67 | 38 | 41.76 | 61.29 | 58.91 | 56.72 |
| 11 | *Russula* | 161 | 138 | 75.82 | 86.79 | 86.25 | 85.71 |
| 12 | *Suillus* | 100 | 94 | 78.99 | 83.16 | 88.26 | 94.00 |
| - | - | 1436 | 1205 | 83.91 | 83.91 | 83.91 | 83.91 |

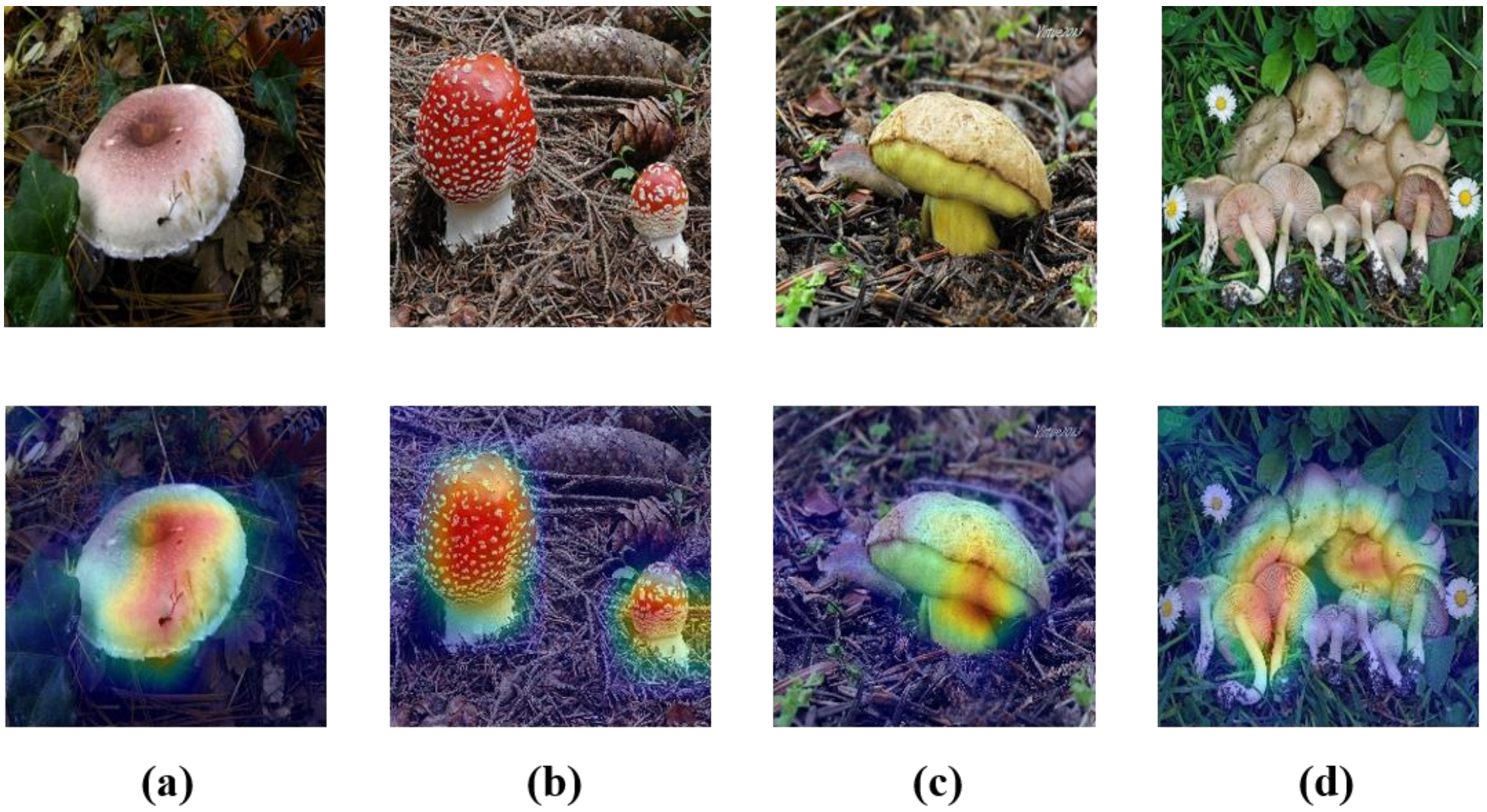

**Figure 11.** Heat map of several mushroom species

### 3.3. Insight into the attention mechanism

In order to explore the impact of different attention strategies on the performance of the model and the focus on images among various layers of the model, we selected 8 different attention strategies, as shown in Figure 12 (a): Model-1 adds an SE module after the last convolution layer, and Model-2 adds an SE module after the first and last convolution layers, Model-3 adds an SE module after the first convolution layer, Model-4 adds two se modules after the first convolution layer, an SE module after the last convolution layer, Model-5 adds an ECANet module after the last convolution layer,

Model-6 adds an SE module after the first convolution layer and an ECANet module after the last convolution layer, Model-7 adds two se modules after the first convolution layer. Finally, the model we proposed adds two se modules after the first convolution layer and an ECANet module after the last convolution layer. Eight models were trained on local data sets, and the training results are shown in Table 6.

For the focus of each layer of the model we proposed, as shown in Figure 12 (b), we selected four species of mushrooms on the local dataset and the open dataset respectively, and visualized each layer of the model. It is obvious that with the deepening of the model level, the focus of the model on the image has gradually narrowed from the initial large-scale attention to the main features of mushrooms, such as *Agaricus* in the open dataset. From the model visualization results of the last layer, it can be seen that the model focuses most of its attention on the mushroom cap, indicating that the model has successfully identified the feature of mushrooms.

The experimental results are shown in Table 3-8. Different attention strategies have a large impact on the performance of the model. From the results from Model 1 to Model 7 and our proposed model, our model has an accuracy of 73.8% in the validation set, second only to Model 6. In the test set, our model has an accuracy rate of 77.4% and a loss rate of 1.73, which is significantly better than other models and achieves the best results.

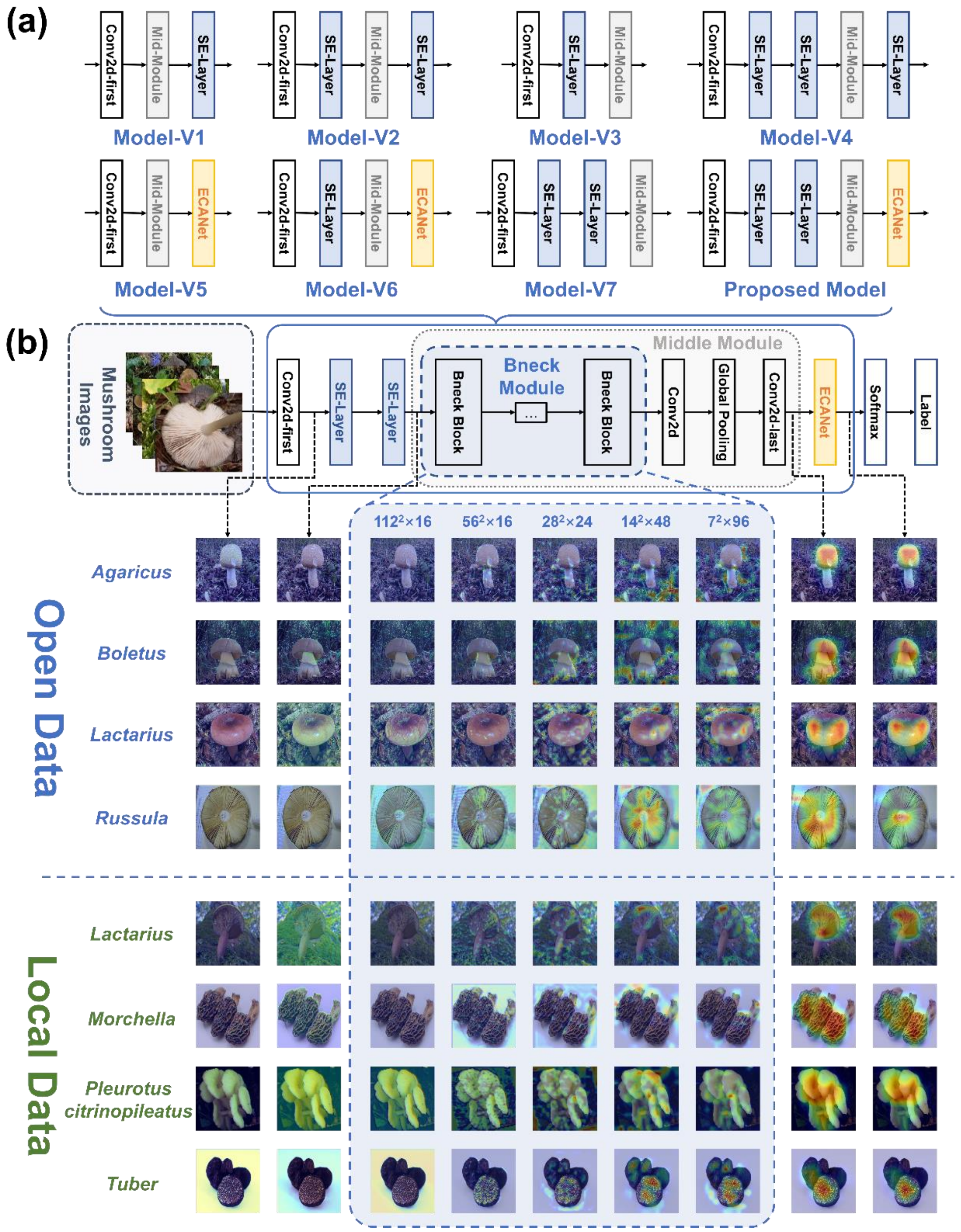

**Figure 12.** Different attention strategies in the model

It can be seen from model 4 and model 7 that model 4 adds an SE module after the last convolutional layer, and its accuracy rate reaches 76.6%, which is 6.1% higher than that of model 7, indicating that adding a SE module at the end has a good effect on the model. The performance has a positive effect, and ECANet retains more information than SE module, and there is no information loss after dimensionality reduction. The last layer of SE module is replaced by ECANet module. The accuracy rate of our proposed model reaches 77.4%, Best results were achieved in models 1-7.

**Table 6.** Accuracy and loss of different models on local dataset

| Models | Val accuracy | Val loss | Test accuracy | Test loss |
|---|---|---|---|---|
| **Model-1** | **71.7%** | **1.20** | **73.9%** | **1.01** |
| **Model-2** | **69.5%** | **1.90** | **70.3%** | **1.77** |
| **Model-3** | **59.6%** | **1.67** | **64.5%** | **1.39** |
| **Model-4** | **72.6%** | **1.98** | **76.6%** | **1.74** |
| **Model-5** | **70.1%** | **1.14** | **72.5%** | **1.07** |
| **Model-6** | **74.4%** | **1.77** | **77.0%** | **1.49** |
| **Model-7** | **67.3%** | **1.61** | **70.5%** | **2.35** |
| **Proposed Model** | **73.8%** | **1.96** | **77.4%** | **1.73** |

## 3.4. Insight into the biometric image classification: through genetic distance

### 3.4.1. Image distance generation based on genetic distance

Traditional image recognition is difficult to measure the distance between images, and the image distance calculated by a given algorithm is difficult to have a correct meaning in real situations, especially in the field of biometrics, the difference measurement between images is difficult to conform to biological semantics. Therefore, in this study, the genetic distance is used as the representation space, and the genetic distance between the mushroom species in the dataset is used as the embedding of the genetic distance representation space. This study uses a convolutional neural network to extract image features and map the image features to the feature space of a fully connected layer (Fc). The feature space and representation space are connected by softmax or MSE function, that is, the Embedding of genetic distance. Through learning on the training set, the model acquires the ability to generate genetic distances from image features, that is, to obtain image distances with biological genetic information through feature extraction (Figure 13a).

This study uses different activation functions to learn genetic distance, including MSE-sum, MSE-mean, and Softmax. The prediction results and errors of genetic distance are shown in Figure 3-8. It can be found that MSE-sum, MSE-mean, the overall prediction accuracy of the genetic distance of the model is high, the average error of the genetic distance between species is about 0.01~0.1, and the diagonal elements and the species with farther genetic distance (*Cantharellus cibarius*) prediction ability decreases, and the error increases to between 0.3 and 0.5. Using the Softmax activation function will greatly enlarge the distance between different species, resulting in the loss of the overall predictive ability of the genetic distance, and the overall predicted value is far beyond the range of 0 to 1. It shows that in the task of image recognition, the Softmax activation function will over-amplify the distance in the high-dimensional space of image feature mapping.

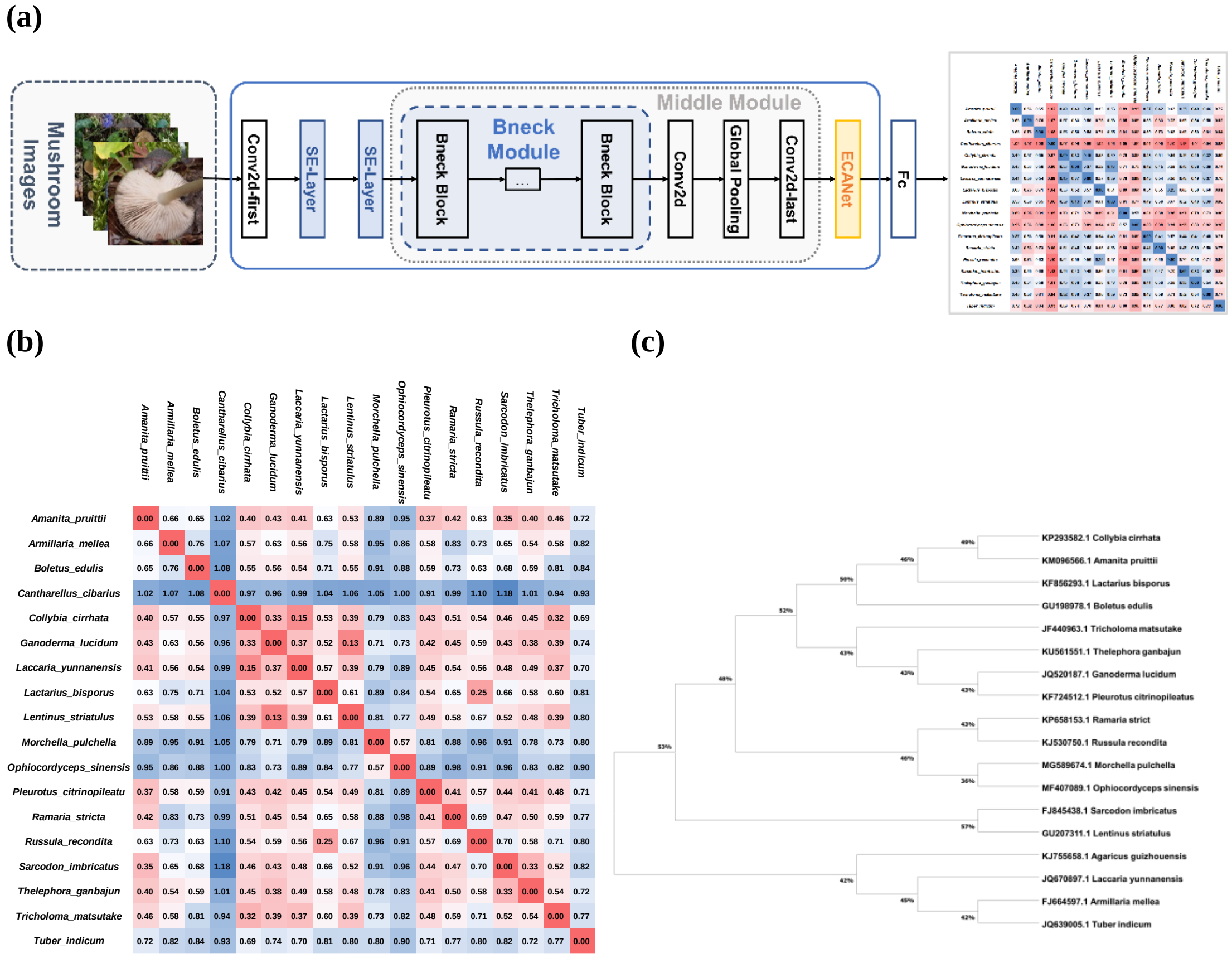

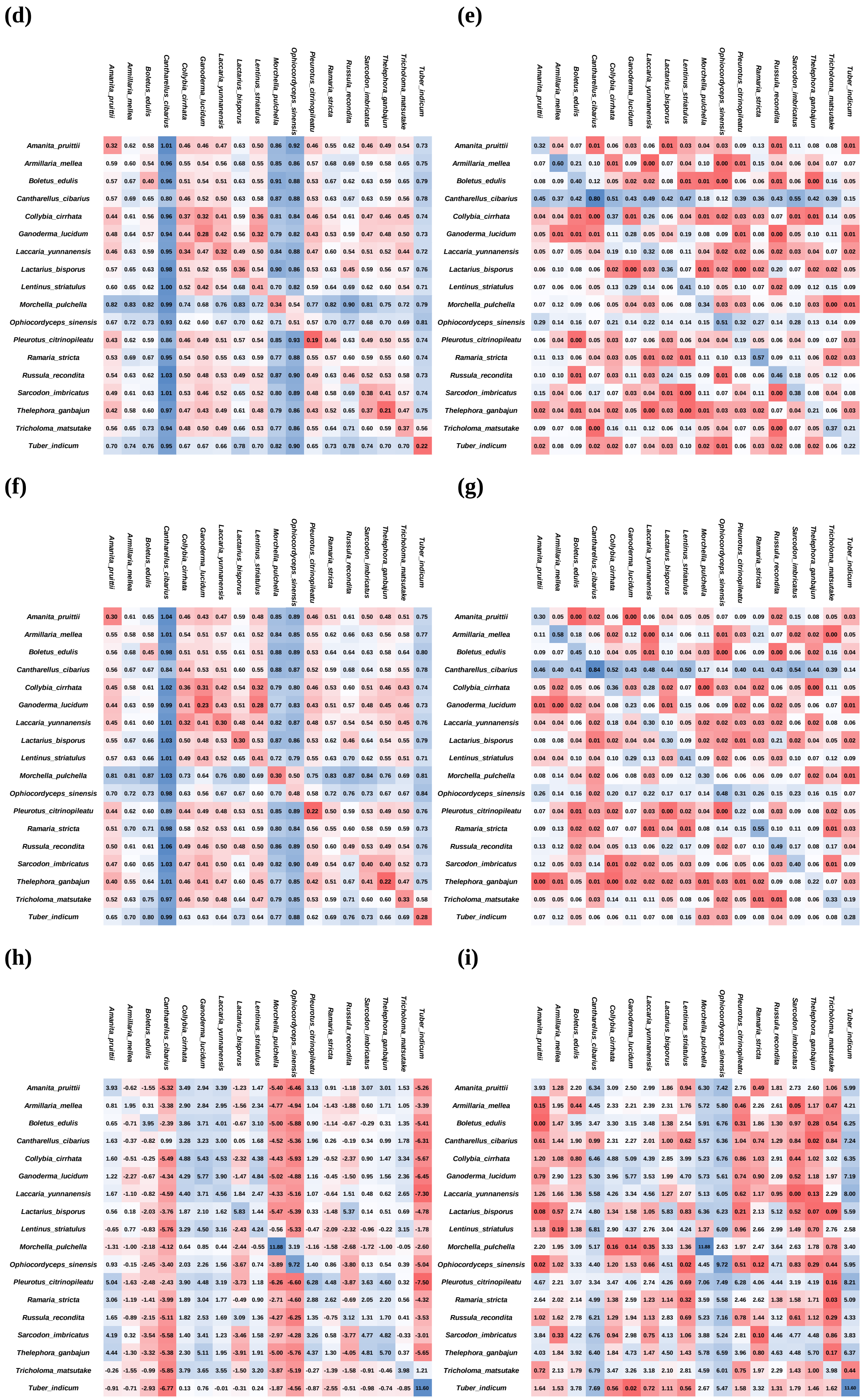

**Figure 13** Schematic diagram of image distance generation based on genetic distance

(b) Genetic distance Ground truth, corresponding to (c) phylogenetic tree, and (d) MSE-sum, (f) MSE-mean, (h) Softmax Predicted genetic distance and error of (e) MSE-sum, (g) MSE-mean, (i) Softmax

### 3.4.2. Improved Image Distance Generation for Genetic Distance Diagonal Element

Based on the above results, this study combines the characteristics of the MSE activation function and Softmax, and uses the elements with a diagonal of -1 to enlarge the distance between the species itself and itself, so as to reduce the error of the model's diagonal element prediction. The genetic distance and error predicted by the model are shown in Figure 14. It can be seen that changing the diagonal elements has significantly improved the overestimation of most of the diagonal elements, and most of the diagonal elements are identified as below 0, which is different from other genetic distances, which is convenient for subsequent species identification tasks. For species other than Cantharellus cibarius, the genetic distance prediction error was at a low level of about 0.01~0.1, indicating that improving the genetic distance diagonal element would not cause the model to lose its ability to estimate the overall genetic distance.

**(a)** **(b)**

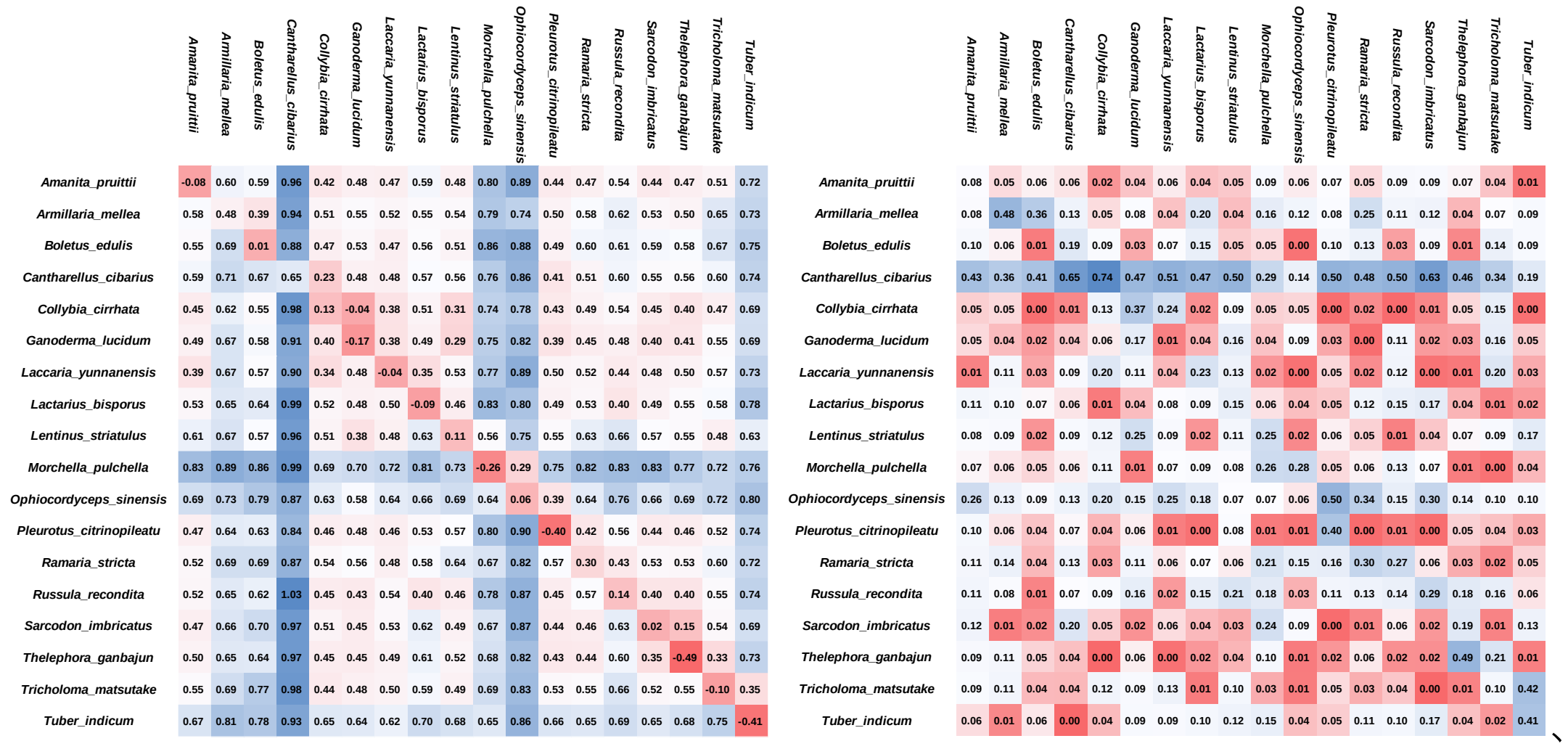

**Figure 14** (a) Genetic distance and (b) error of MSE-sum prediction of improved diagonal elements

### 3.4.3. Species identification based on genetic distance

In this study, the distance calculation is performed between the Embedding of the genetic distance predicted by the model and the ground truth of the genetic distance, including cosine distance, Euclidean distance, etc. The obtained distance is the Label Embedding, which can be mapped to the Label space through the Label Embedding to complete mushroom image classification.

In order to study the problem of species identification based on genetic distance, this study firstly studied the influence of the normalization of genetic distance on the identification accuracy. The test accuracies of Proposed Model-softmax-normalized, Proposed Model-MSE-A-normalized and Proposed Model-MSE-S-normalized after normalization of genetic distance identification are 66.1%, 57.9% and 61.5%. Among them, the accuracy rate of Proposed Model-MAE-normalized is the lowest at 22.4%, and it is almost impossible to perform category recognition tasks. Compared with the identification on the original genetic distance, the accuracy of the Proposed Model-softmax is 74.3%, indicating that normalization will seriously affect the estimation of the genetic distance by the Softmax activation function. For the MSE function, the accuracy rates of Proposed Model-MSE-A and Proposed Model-MSE-S are 57.9% and 61.5%, with almost no change, so the normalization process uses the model of the MSE activation function to estimate the influence of genetic distance Not much. When the model reduced the Cantharellus cibarius species, the accuracy of the Proposed Model-MSE-17 improved slightly, indicating that species with greater genetic distances adversely affected the predictive power of the model. When the model uses the strategy of setting the diagonal element to -1, the Proposed Model-MSE-(-1) model achieves a validation accuracy of 68.1% and a test accuracy of 72.5%, which is close to the best using the Softmax activation function. Accuracy. It shows that the diagonal elements of the improved genetic distance can significantly improve the model's ability to identify species without losing the predictive ability of the model for genetic distance, and it is easier to distinguish different species in the representation space of genetic distance. As shown in Table 7.

**Table 7.** Accuracy and loss of different models on local dataset

| Models | Val accuracy | Val loss | Test accuracy | Test loss |
|---|---|---|---|---|
| **Proposed Model-softmax-normalized** | **57.4%** | **24.68** | **66.1%** | **24.46** |
| Proposed Model-MSE-A-normalized | 51.8% | 20.70 | 57.9% | 20.31 |
| Proposed Model-MSE-S-normalized | 51.4% | 44.83 | 61.5% | 43.71 |
| Proposed Model-MAE-normalized | 20.3% | 42.20 | 22.4% | 41.80 |
| **Proposed Model-softmax** | **69.9%** | **32.95** | **74.3%** | **33.60** |
| Proposed Model-MSE-A | 51.8% | 0.21 | 57.9% | 0.20 |
| **Proposed Model-MSE-S** | **51.4%** | **44.82** | **61.5%** | **43.71** |
| Proposed Model-MSE-17 | 54.8% | 5.23 | 63.7% | 67.61 |
| **MobileNetV3-MSE-(-1)** | **65.9%** | **27.13** | **67.5%** | **99.05** |
| **Proposed Model-MSE-(-1)** | **68.1%** | **22.36** | **72.5%** | **98.96** |

The confusion matrix and ROC curve of the MSE-mean-(-1) and MSE-sum-(-1) models using the Softmax activation function and the improved diagonal elements in this study are shown in Figure 3-9. It can be found that the ROC of the model using the Softmax activation function is relatively high, but the model using the MSE activation function can also achieve an average AUC value above 0.8.

**(a)**

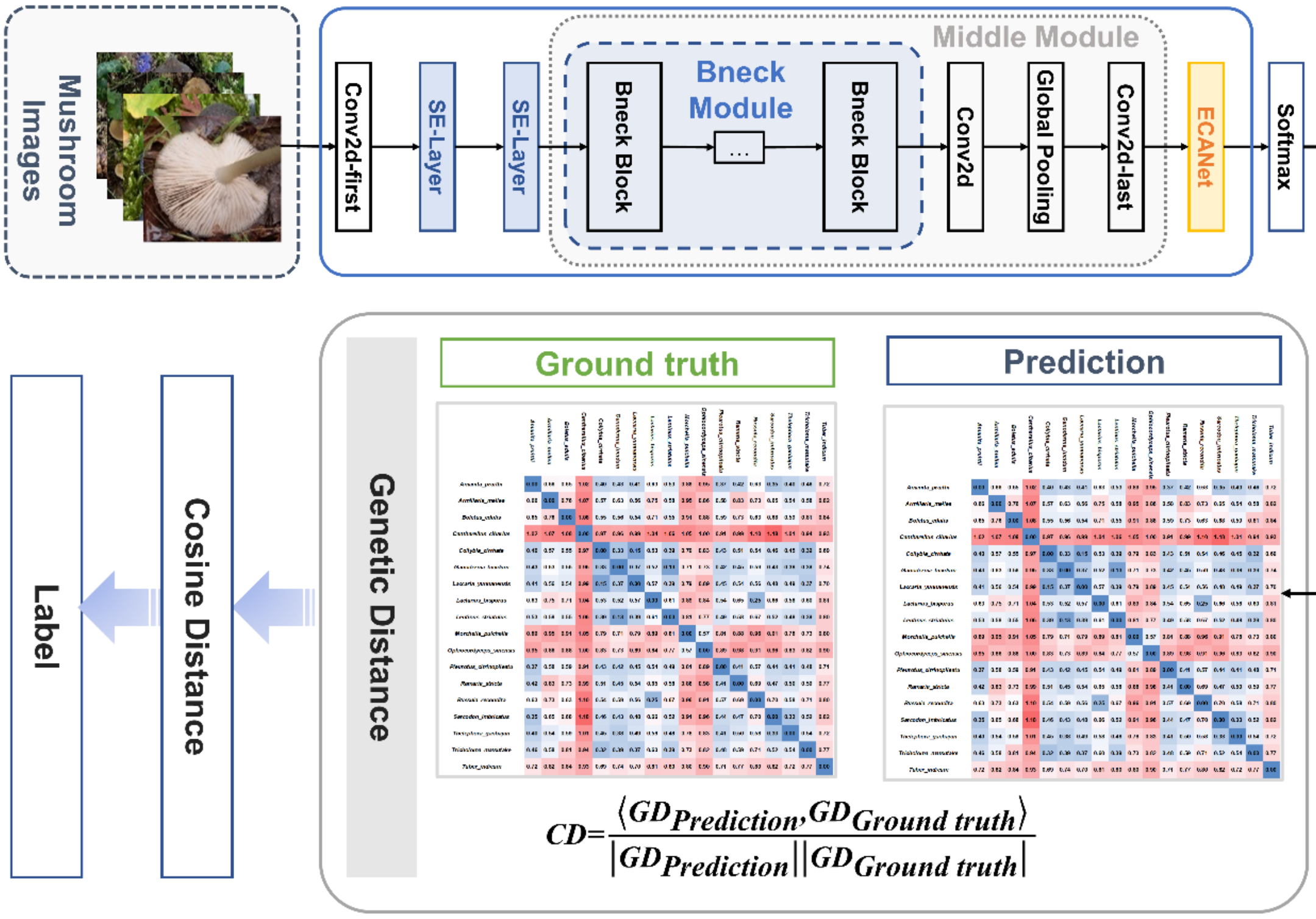

**(b)** **(c)**

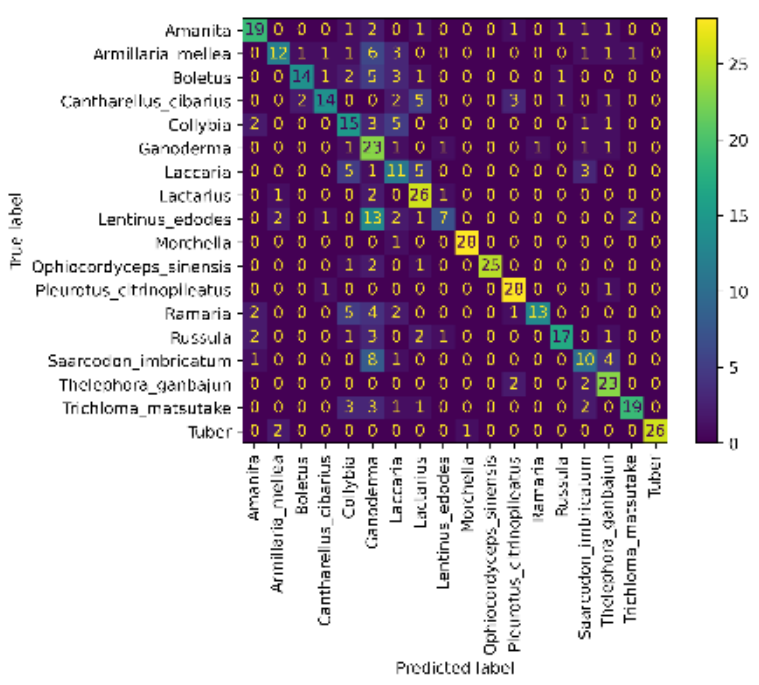

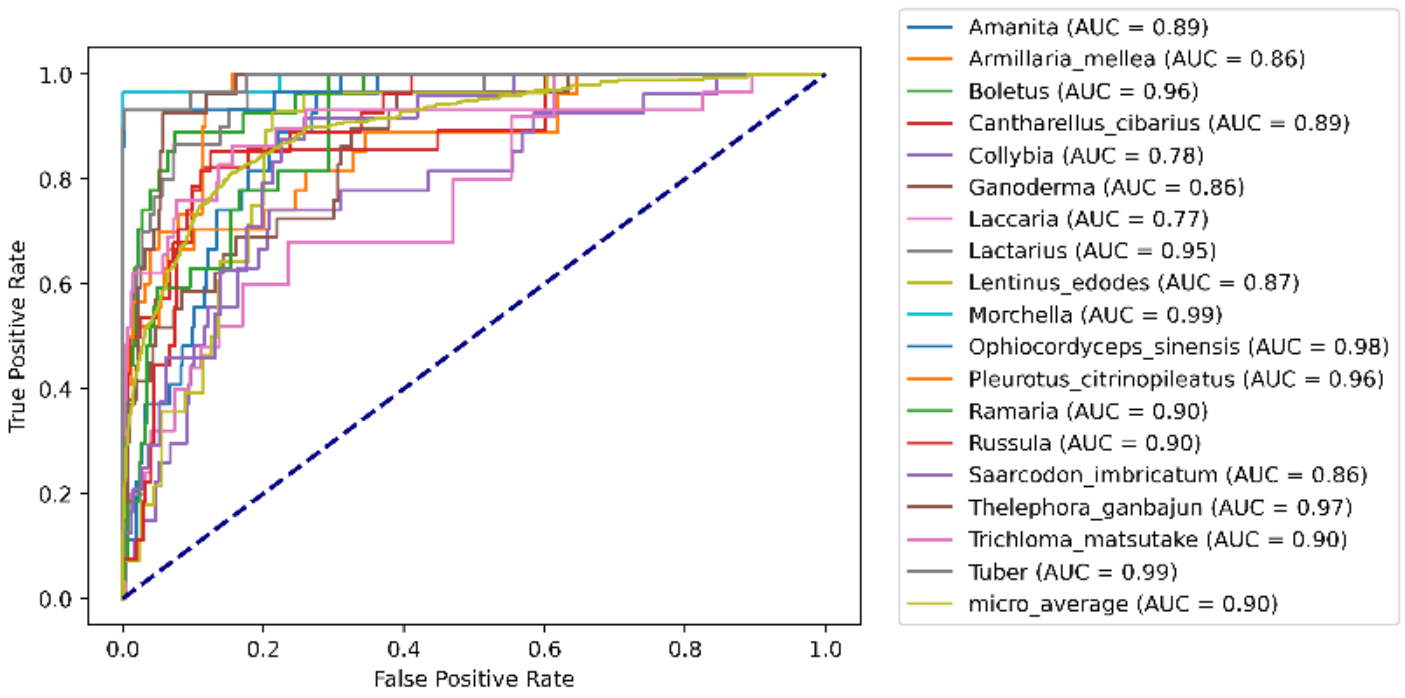

**(d)** **(e)**

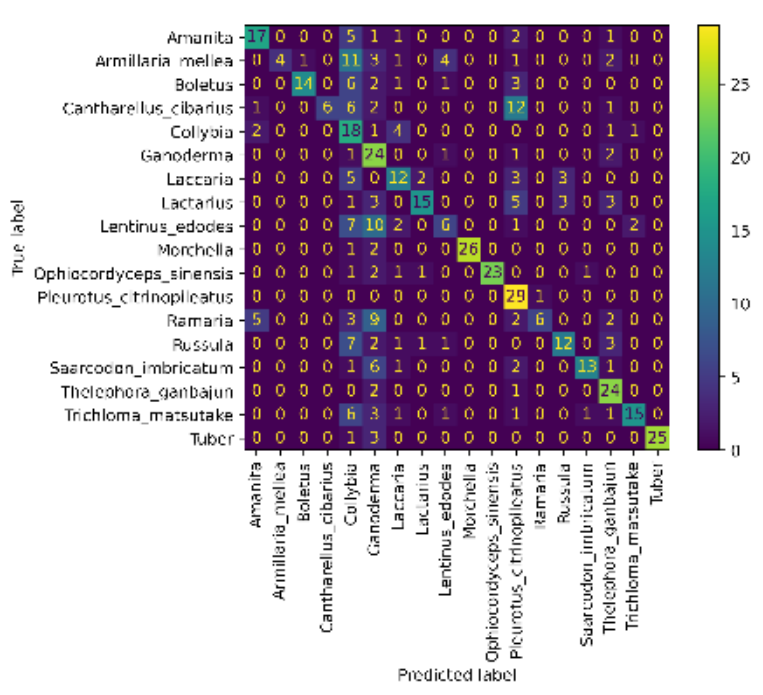

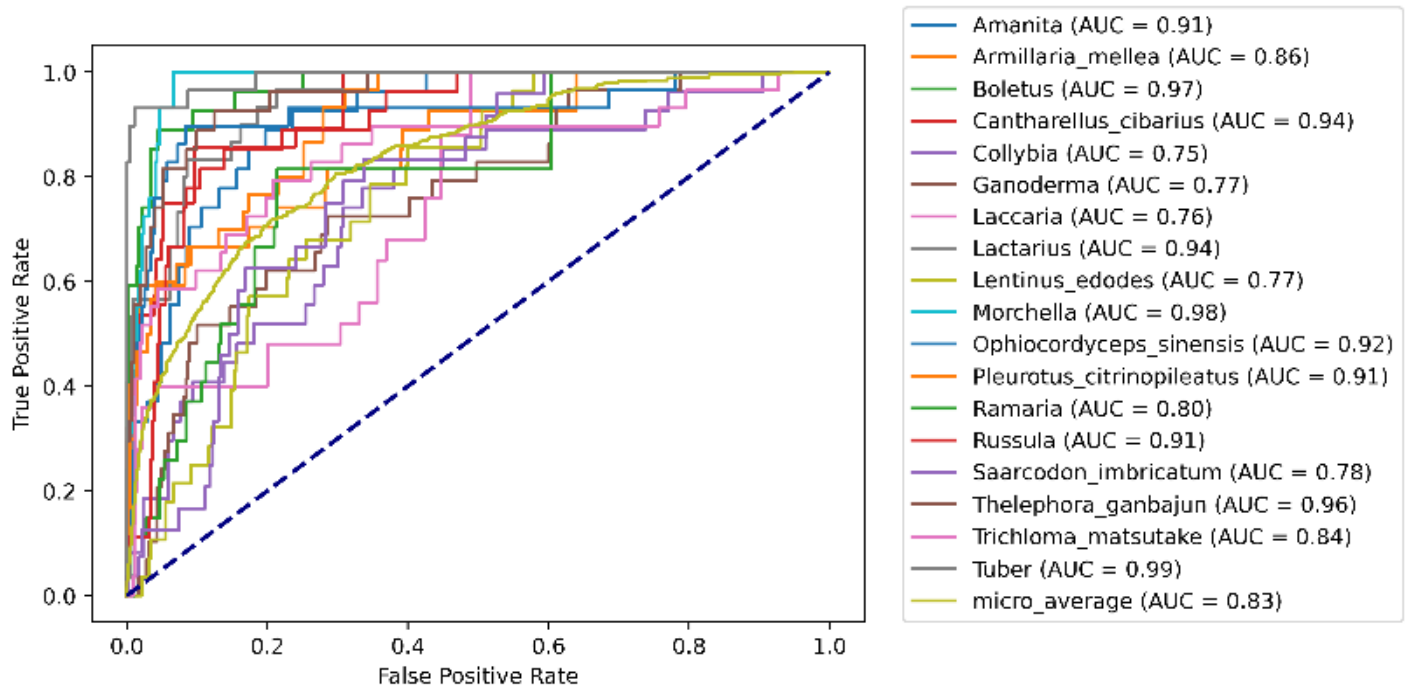

**(f)** **(g)**

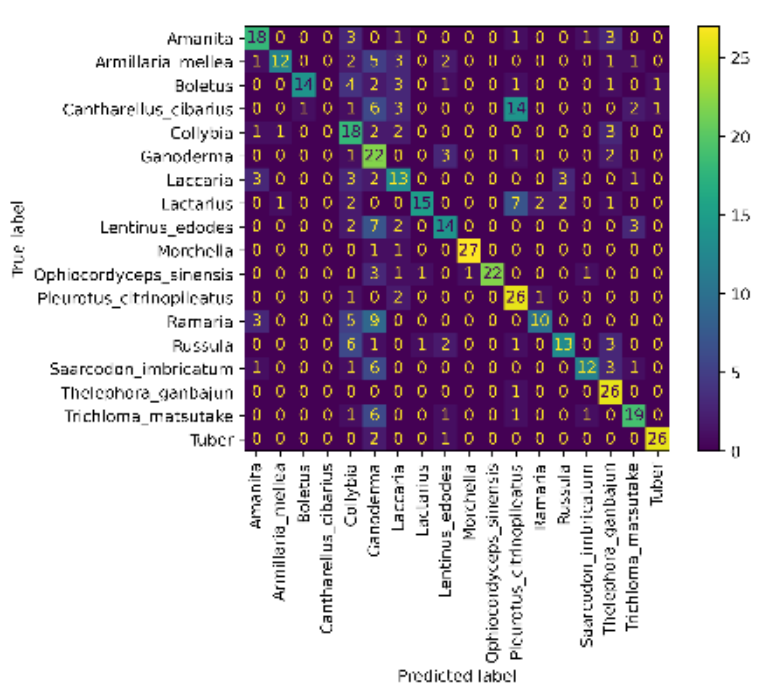

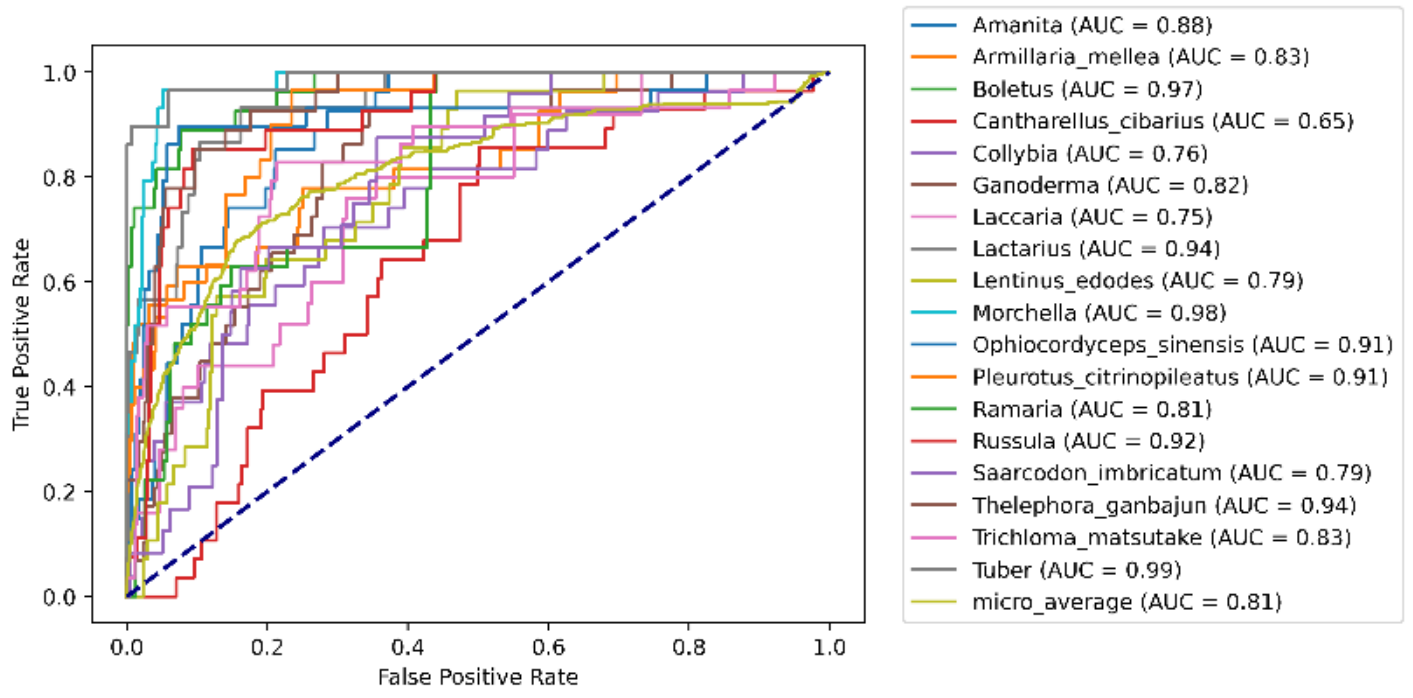

**Figure 15** (a) Species recognition based on genetic distance, and the confusion matrices of different activation functions (b) Softmax, (d) MSE-mean-(-1), (f) MSE-sum-(-1) and (c) Softmax, (e) ROC curves of MSE-mean-(-1), (g) MSE-sum-(-1)

## 4. Conclusions

Identification and classification of various mushroom species by deep convolutional neural network can increase efficiency in production efficiency effectively and efficiently. However, there are many kinds of mushrooms and the interference of living environment is large. Mushroom recognition itself is a difficult task, and the ordinary convolutional neural network model usually has a large number of parameters and cannot be deployed on the mobile terminal, so in this paper, we proposed a novel network architecture called MushroomNet, which uses the lightweight network mobilenetv3 as the backbone model. In order to improve the performance of the performance model on a small sample dataset, we load the parameters and weights trained from ImageNet through the technology of migration learning, and then select the best attention mechanism and attention structure to be embedded in our proposed model. We train and test our model on local dataset and public dataset, and it shows excellent performance in mushroom recognition task. Furthermore, we train our model by add the genetic distance as the representation space and the genetic distance between two mushroom species as the embedding of image distance. Experiments show that our model shows small error rates in predicting genetic distance at the expense of certain recognition accuracy, our proposed model is suitable for mobile terminal to identify and classify mushrooms in the field, it can also be used as an efficient classification system for rapid batch identification of mushrooms.

**Acknowledgments**

This work was supported by the National Undergraduate Training Programs for Innovation and Entrepreneurship (202110022063), the Projects of Science and Technology of Programs of Guizhou Province ([2019]-4007, [2018-4002]), and the Fifth Batch of "Thousands of Innovative and Entrepreneurial Talents" in Guizhou Province.

**Data availability**

Our multi-perspective mushroom datasets are available from

https://www.kaggle.com/datasets/jiewenxiao/mushroom-classification.

## Code availability

Our codes are available from https://github.com/set-path/mushroom-recognization.

## Contributions

The work was conceptualized by W. L., J. X. and J. W. In details, J. X., W. L., design and built the model from J. W. and Y. Y. The local mushroom dataset was constructed by W. L., C. Z. and J. X. J. X. proposed the genetic information meditating methods. W. L., Y. H., Z. G. and J. W. wrote the software. W. L. and J. X. designed the experiments, analyzed and visualized the data, wrote the original draft. J. W., W. L., J. X., C. Z., Y. H. and Z. G. reviewed and edited the manuscript. J. W. and Y. Y. supervised the project.